\documentclass[times,twocolumn,final,nopreprintline]{elsarticle}

\usepackage{medima}
\usepackage{framed,multirow}

\usepackage{algorithmic}
\usepackage{algorithm}
\usepackage{array}
\usepackage[caption=false,font=normalsize,labelfont=sf,textfont=sf]{subfig}
\usepackage{textcomp}
\usepackage{stfloats}
\usepackage{url}
\usepackage{verbatim}
\usepackage{caption}
\usepackage{graphicx}
\usepackage{amsmath,amssymb,amsfonts}
\usepackage{epsfig}
\usepackage{dsfont}
\usepackage{color}
\usepackage{boldline,multirow }
\usepackage{siunitx}
\usepackage{booktabs}
\usepackage{soul}
\usepackage{cases}
\usepackage{hyperref}
\usepackage{bbding}
\usepackage{makecell}
\usepackage{placeins}

\usepackage{graphicx}   
\usepackage{caption}    
\usepackage{adjustbox}  

\usepackage[dvipsnames, table]{xcolor}  
\usepackage{booktabs}   

\definecolor{first}{RGB} {189, 230, 205}
\definecolor{second}{RGB}{228, 238, 188}

\let\oldcolorbox\colorbox

\newcommand{\colorboxcontent}[2]{%
    \ifstrequal{#1}{first}{\textbf{#2}}{%
        \ifstrequal{#1}{second}{\underline{#2}}{#2}%
    }%
}

\RenewDocumentCommand{\colorbox}{o m m}{%
    \IfNoValueTF{#1}
        {\oldcolorbox{#2}{\colorboxcontent{#2}{#3}}}
        {\oldcolorbox[#1]{#2}{\colorboxcontent{#2}{#3}}}%
}

\journal{Medical Image Analysis}

\begin{document}

\verso{Z. Xu, \textit{et~al.}}

\begin{frontmatter}

\title{PathPilot: An Adaptive Question-Conditioned Multi-Magnification Navigation Agent for Whole-Slide Image Analysis}

\author[inst1,inst2]{Zhengyang \snm{Xu}}
\ead{zhengyang.xu@tum.de}

\author[inst1,inst2,inst3]{Han \snm{Li}}

\author[inst1,inst2]{Jingsong \snm{Liu}}

\author[inst5]{Ayako \snm{Ura}}

\author[inst6]{Shihui \snm{Zu}}

\author[inst4]{Shaohua Kevin \snm{Zhou}}

\author[inst2,inst3]{Nassir \snm{Navab}}

\author[inst1,inst2,inst7]{Peter \snm{Schüffler}}
\ead{peter.schueffler@tum.de}

\address[inst1]{Institute of Pathology, Technical University of Munich, Germany}
\address[inst2]{Munich Center for Machine Learning, Munich, Germany}
\address[inst3]{Chair of Computer Aided Medical Procedures, Technical University of Munich, Germany}
\address[inst5]{
Department of Human Pathology, Juntendo University Graduate School of Medicine, Tokyo, Japan}
\address[inst6]{School of Biomedical Engineering, Faculty of Medicine, Dalian University of Technology, China}
\address[inst4]{University of Science and Technology of China, China}
\address[inst7]{Munich Data Science Institute, Technical University of Munich, Germany}

\begin{abstract}
The diagnostic capability of Whole-Slide Image (WSI) agents relies on effective slide navigation to identify diagnostically relevant regions and efficiently integrate evidence across the slide. While in clinical practice, pathologists dynamically examine slides across multiple scales, integrating global and cellular evidence in a sequential manner, existing AI navigation approaches mostly operate at a single fixed magnification or rely on predefined magnification traversal. This mismatch prevents existing methods from \textbf{modeling cross-magnification interactions} and \textbf{adaptive magnification selection} inherent to real diagnostic workflows.
We propose \textbf{PathPilot}, a clinically consistent multi-magnification WSI navigation agent that explicitly models cross-magnification localization and evidence-adaptive exploration. Specifically, we introduce a Cross-Magnification Localization Tool (CrossMagLoc) that learns magnification-specific diagnostic localization from pathologist viewing behavior and integrates information from adjacent magnifications to identify the diagnostically most relevant region within the current ROI, thereby providing a diagnostically informed starting point for subsequent question-conditioned exploration. We further introduce an Evidence-Adaptive Region Search Tool (EviRegionSearch) that starts question-conditioned exploration from this region, progressively accumulating question-relevant evidence and adaptively determining the next navigation action according to the remaining evidence requirement. Their iterative interaction enables PathPilot to sequentially explore across ROIs and magnifications until sufficient evidence is acquired.
Extensive QA evaluations across five cohorts demonstrate consistent superiority over competing SOTA methods, with GACC improvements of \textbf{8.46\%} and \textbf{9.03\%} on inter/external cohorts, respectively, validating question-conditioned evidence acquisition across regions and magnifications that mirrors the sequential, knowledge-driven navigation of pathologists. Code will be public upon acceptance.
\end{abstract}

\end{frontmatter} 

\begin{figure}[th]
  \centering
\includegraphics[width=0.5\textwidth]{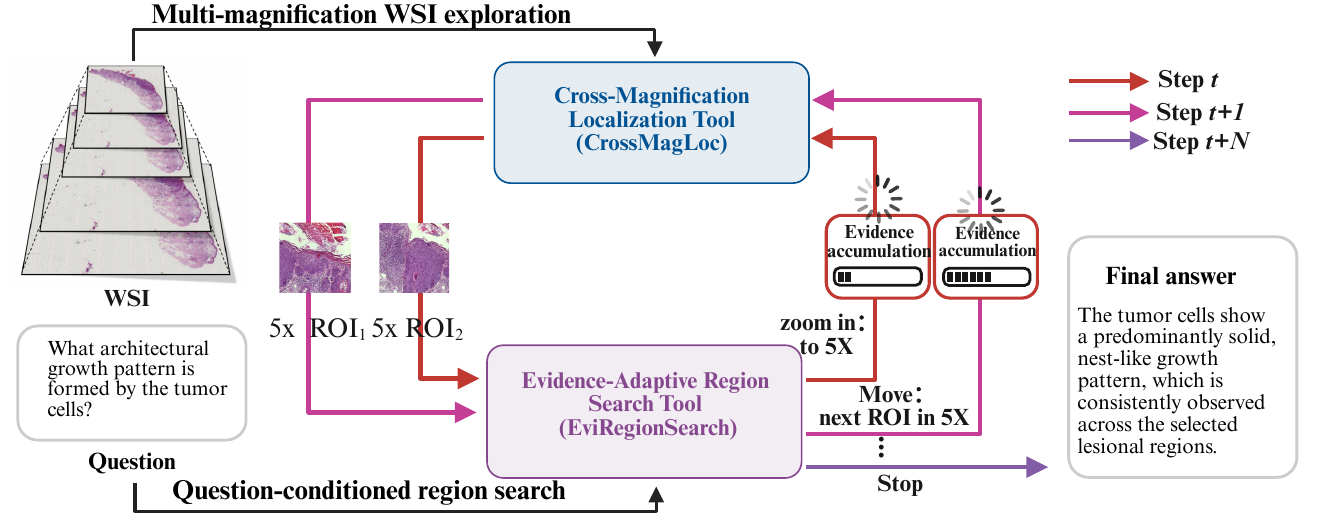}
\caption{
\textbf{Overview of PathPilot for adaptive WSI exploration.} For each navigation step (e.g., Step $t$), CrossMagLoc  identifies the diagnostically most relevant region within the current ROI, while EviRegionSearch accumulates question-relevant evidence and adaptively selects the next navigation action (e.g., Step $t+1$). Their iterative interaction enables sequential movement across ROIs and magnifications until evidence is sufficient (e.g., Step $t+N$), where \textbf{Stop} is selected and the final answer is generated.}
 \label{fig1}
\end{figure}

\section{Introduction}\label{sec:intro}

Recent advances in artificial intelligence have enabled the development of increasingly capable agents for Whole-Slide Image (WSI) analysis, with the potential to perform iterative visual exploration, evidence acquisition, and diagnostic reasoning over gigapixel pathology slides \cite{ghezloo2025pathfinder,chen2025pathagent,yang2026pathnavigate,wang2026pathology}. Effectively analyzing WSIs remains challenging because diagnostically relevant information is sparsely distributed across extremely large spatial domains and may appear at different magnification levels within a WSI \cite{van2021hooknet,bilal2023aggregation}. In routine diagnosis, pathologists therefore do not inspect the slide uniformly. Instead, they first identify regions that are potentially diagnostically informative across magnification levels, and then selectively examine different regions according to the specific pathological question under consideration \cite{van2021deep,nan2025deep,srinidhi2022self}. Such a sequential and selective diagnostic workflow makes WSI analysis particularly well suited to agent-based approaches, where an agent can iteratively explore the slide and gather visual evidence as needed.

Existing AI navigation approaches improve WSI analysis by modeling spatial exploration and identifying diagnostically relevant regions  \cite{raciti2023clinical,pantanowitz2020accuracy}. Compared with conventional patch-based approaches that treat WSIs largely as unordered collections of independently processed patches \cite{ilse2018attention,zhao2022setmil,ding2023multi}, navigation-based methods introduce structured spatial exploration, allowing diagnostically informative regions to be selectively examined across the slide \cite{nan2025deep,ghezloo2025pathfinder,sun2025cpathagent}. More recently, navigation has been incorporated into pathology agents \cite{chen2025pathagent,yang2026pathnavigate}, extending such spatial exploration toward question-conditioned WSI analysis.

However, existing navigation approaches still face two key limitations. First, \textbf{fixed-magnification localization}: many methods operate at a single magnification, preventing them from identifying diagnostically relevant regions across different magnification levels. Second, \textbf{exploration is not sufficiently flexible}: although existing agents can autonomously select regions to inspect, they have limited ability to adjust their search across regions and magnification levels according to the information being sought. Both limitations stand in contrast to clinical practice, where pathologists identify diagnostically relevant regions across magnifications and continuously adapt their exploration according to the evidence still needed, integrating tissue-level architectural context with cellular-level details \cite{thandiackal2022differentiable,agarwal2021survival,ghezloo2022analysis,lee2024convolutional}.

To address these limitations, we propose PathPilot, an adaptive question-conditioned Multi-Magnification WSI Navigation Agent with two sequentially coupled components (Fig.~\ref{fig1}). The \textbf{Cross-Magnification Localization Tool (CrossMagLoc)} performs question-agnostic diagnostic localization within the current Region of Interest (ROI), identifying the diagnostically most relevant region independently of the current question. The \textbf{Evidence-Adaptive Region Search Tool (EviRegionSearch)} then starts its question-conditioned exploration from this region according to the current question and accumulated evidence.  As evidence progressively accumulates, EviRegionSearch identifies the remaining evidence requirement and selects the next navigation action accordingly. Specifically:

The \textbf{CrossMagLoc} (Sec.~\ref{sec:CrossMagLoc}) learns magnification-specific diagnostic localization from pathologist viewing behavior. Given the current ROI, it generates a question-agnostic diagnostic relevance heatmap and identifies the diagnostically most relevant region, providing a starting point for subsequent question-conditioned exploration. To move beyond fixed-magnification localization, we introduce learnable magnification tokens that capture the diagnostic focus at each magnification. Guided by these tokens, CrossMagLoc incorporates information from the current and adjacent magnification levels to improve navigation performance.

The \textbf{EviRegionSearch} (Sec.~\ref{sec:EviRegionSearch}), starting from the most diagnostically relevant ROI identified by the question-agnostic heatmap generated by CrossMagLoc, performs iterative, question-conditioned exploration according to the current question and accumulated question-relevant evidence. At each step, it examines the ROI selected through the previous action, updates the accumulated question-relevant evidence, and dynamically selects which candidate ROI and magnification level should be examined next to continue evidence accumulation. This iterative process enables movement to another region within the previous step's ROI at the current magnification (Move), closer inspection at a higher magnification (Zoom In), or a return to a lower magnification (Zoom Out), and terminates when sufficient evidence has been accumulated to answer the question (Stop).

Obviously, \textbf{evaluating diagnostic navigation in WSIs is inherently challenging} because there is no unique ground-truth navigation behavior. Different pathologists may inspect different yet equally diagnostic regions, while their viewing patterns can also vary substantially with experience and individual practice \cite{ghezloo2022analysis, nan2025deep,chakraborty2025measuring}. Therefore, neither reproducing a particular pathologist's viewing trajectory nor matching a single set of attended regions can fully characterize navigation quality. To address this inherent evaluation challenge, \textbf{we assess PathPilot from multiple complementary perspectives}: whether navigation ultimately supports accurate WSI analysis, whether the localized regions agree with pathologist prioritization, whether they cover objectively annotated pathological regions, and whether their diagnostic relevance is confirmed by independent pathologists.

Following this evaluation strategy, we conduct experiments on \textbf{five open-ended question-answer (QA) cohorts} spanning skin, liver, and uterine pathology, where answers are generated freely rather than selected from predefined options, together with \textbf{three navigation datasets}  (Sec.~\ref{sec:datasets}). Specifically, we first evaluate the overall WSI analysis performance of the complete PathPilot framework through open-ended QA under both internal and external validation settings (Sec.~\ref{sec:qa_exp}). We then independently benchmark the diagnostic localization provided by CrossMagLoc against existing navigation methods, assessing both alignment with pathologist-derived region prioritization and coverage of annotated tumor regions across magnifications (Sec.~\ref{sec:nav_exp}). The contribution of EviRegionSearch is further examined through controlled ablation studies (Sec.~\ref{sec:eviregionsearch_eval}). Moreover, qualitative analyses, including independent pathologist assessment and a representative PathPilot workflow, provide complementary evaluation of the diagnostic relevance of the selected regions and the adaptive exploration behavior of PathPilot (Sec.~\ref{sec:qualitative}).

Across all five open-ended QA cohorts, PathPilot achieves the highest overall GACC, \textbf{outperforming SOTA methods by an average of 8.46\% and 9.03\%} on the internal and external cohorts, respectively. For navigation, CrossMagLoc achieves the strongest agreement with pathologist region prioritization, \textbf{improving region-ranking correlation from 0.56 to 0.70} over the best baseline, while \textbf{covering 85--100\% of annotated tumor regions across magnifications}. Independent pathologist review further confirms the diagnostic relevance of the localized regions. 
Together, these complementary evaluations support the effectiveness of PathPilot for both WSI analysis and navigation.
Our results suggest superiority of PathPilot's dynamic region and resolution selection over conventional fixed sampling at static magnifications, thus offering \textbf{a new way for adaptive and effective WSI analysis} in computational pathology.

\section{Related Work}
\subsection{Vision-Language Models in Computational Pathology.}

Vision-language models (VLMs) have rapidly advanced multimodal understanding by coupling visual representations with the reasoning and generation capabilities of large language models. General-purpose VLMs, initially developed for broad natural-image and multimodal understanding, have progressively improved visual instruction following and multimodal reasoning, with representative models including Gemma-3  \cite{Kamath2025Gemma3T}, Qwen-VL  \cite{bai2025qwen25vl,bai2025qwen3}, and InternVL  \cite{zhu2025internvl3,wang2025internvl3}. Their strong general visual-language capabilities have provided the foundation for subsequent adaptation to specialized medical domains.

General medical VLMs further adapt these capabilities to heterogeneous medical images and clinical tasks. Representative approaches include LLaVA-Med  \cite{li2023llava}, HuatuoGPT-Vision  \cite{chen2024towardsHuatuoGPT}, Med-Flamingo  \cite{moor2023med}, Lingshu  \cite{xu2025lingshu}, MedGemma  \cite{sellergren2025medgemma}, HealthGPT-Pro  \cite{lin2025healthgptmedicallargevisionlanguage}, and Hulu-Med  \cite{jiang2025hulu}. Importantly, the distinction between general medical and pathology-specialized VLMs does not imply the absence of pathology data in the former: several recent general medical models explicitly incorporate histopathology within their multimodal medical training, including MedGemma, HealthGPT-Pro, and Hulu-Med, while MedGemma 1.5 further extends general medical modeling to histopathology WSIs  \cite{sellergren2026medgemma}. Nevertheless, these models are designed to provide broad medical visual understanding across multiple imaging modalities rather than being specifically optimized for computational pathology.

In parallel, pathology-specialized VLMs have been developed to more explicitly model histological morphology and pathology-specific visual-language associations. Quilt-LLaVA  \cite{seyfioglu2024quilt}, PA-LLaVA  \cite{dai2024pa}, PathGen-LLaVA  \cite{sun2025pathgen}, and Patho-R1  \cite{zhang2026patho} learn from pathology-specific image--text or instruction data and substantially strengthen fine-grained morphological understanding. These models primarily operate on individual pathological images or localized regions. When directly applied to WSI analysis, a downsampled slide thumbnail can therefore be treated as a single visual input; although the vision encoder internally tokenizes the image into patches, the model does not explicitly aggregate independently sampled regions distributed across the gigapixel WSI.

To overcome the limited spatial coverage of single-view models, recent pathology VLMs have extended visual-language modeling toward whole-slide analysis. WSI-VQA  \cite{chen2024wsi}, WSI-LLaVA  \cite{liang2025wsi}, SlideChat  \cite{chen2025slidechat}, and ANTONI-Alpha  \cite{moonemans2025democratizing} construct WSI-level representations by incorporating information from multiple spatial regions or slide-level visual features, enabling broader pathological context to be considered during language reasoning. MedGemma 1.5 similarly introduces explicit support for histopathology WSIs within a general medical VLM  \cite{sellergren2026medgemma}. Compared with single-view models, these approaches substantially improve access to spatially distributed slide information; however, the visual evidence is generally obtained through a predefined patch-sampling or slide-representation procedure rather than being actively acquired according to the evolving evidence requirement of each pathology question.

\subsection{Navigation Agents in Computational Pathology.}

While WSI-level VLMs broaden visual context by aggregating information from a predefined set of slide regions, the visual evidence available for reasoning is largely determined before the question is answered. Navigation-based analysis instead treats WSI understanding as a sequential evidence-acquisition process, in which diagnostically relevant regions are selectively inspected and newly acquired observations can guide subsequent exploration. This paradigm more closely resembles routine pathological review, where pathologists continuously redirect attention across tissue regions and visual scales according to the emerging diagnostic evidence. Such selective exploration is clinically meaningful because it avoids treating all slide regions as equally informative, provides an interpretable trajectory of the evidence supporting a diagnostic decision, and enables visual inspection to be progressively focused on regions most relevant to the current diagnostic task.

Early computational studies began to model selective WSI inspection by learning directly from pathologists' viewing behavior. PathAttFormer  \cite{chakraborty2022predicting} models pathologists' visual attention during WSI examination and predicts diagnostically informative regions from low-resolution slide representations. Building on explicit pathologist viewing behavior, PEAN  \cite{nan2025deep} further learns diagnostically relevant region importance from large-scale eye-tracking records collected during routine WSI review, including gaze locations, viewing magnifications, and diagnostic decisions. More recently, pathology agents have incorporated navigation directly into iterative diagnostic workflows. PathFinder  \cite{ghezloo2025pathfinder} employs a dedicated Navigation Agent that iteratively generates importance maps to identify diagnostically relevant regions and guide patch sampling for subsequent description and diagnosis. PathAgent  \cite{chen2025pathagent} employs an LLM-based Navigator--Perceptor--Executor framework to iteratively localize informative regions, extract morphological observations, and incorporate them into an evolving reasoning trajectory. More recently, PathNavigate  \cite{yang2026pathnavigate} formulates WSI-VQA as a scan--search--readout process, combining low-magnification slide scanning, question-conditioned target selection, and shared slide memory to retrieve sparse high-resolution evidence.

These approaches demonstrate the value of replacing static WSI representations with selective and iterative region exploration, but an important challenge remains in jointly determining \textbf{where to look}, \textbf{at which magnification}, and \textbf{whether additional evidence is still required} for a specific pathology question  \cite{ghezloo2022analysis,lopes2024eye}. Earlier behavior-supervised navigation methods learn diagnostically informative viewing patterns from pathologists, whereas recent navigation agents introduce question-conditioned region exploration and iterative reasoning. However, the relationship between pathologist-derived navigation priors across multiple magnifications and the question-specific evidence accumulated during agent reasoning remains insufficiently modeled. Our work addresses this gap by coupling multi-magnification diagnostic localization learned from pathologist viewing behavior with question-conditioned navigation action selection and evidence reasoning. CrossMagLoc provides question-agnostic heatmaps for diagnostic region prioritization, while EviRegionSearch selects navigation actions according to the current question, accumulated evidence, and remaining evidence requirement.

\begin{figure*}[tbh]
  \centering
\includegraphics[width=1.0\textwidth]{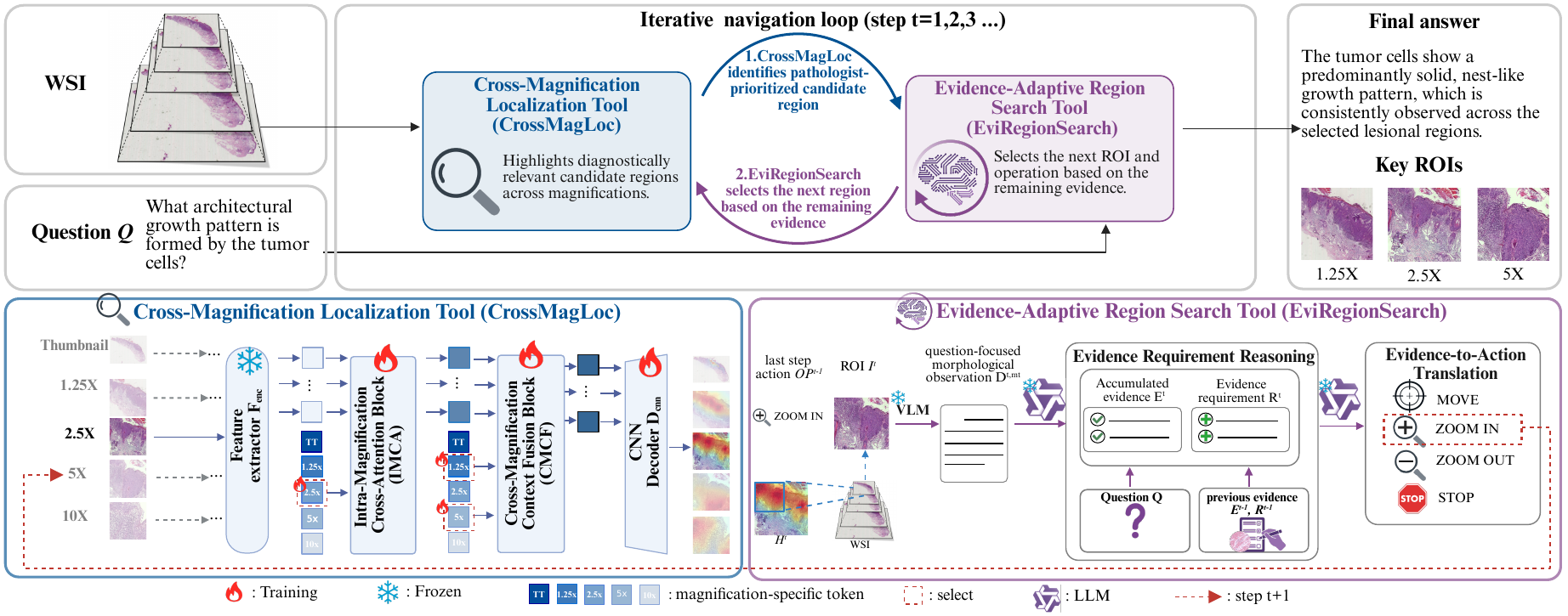}
\caption{\textbf{Overview of PathPilot and its two complementary navigation components. }
Given a WSI and pathology question, PathPilot starts from the thumbnail and iteratively alternates between CrossMagLoc and EviRegionSearch until sufficient evidence has been accumulated for answering the question. CrossMagLoc performs \textbf{question-agnostic localization} of diagnostically relevant candidate ROIs at each magnification by combining magnification-specific representations with spatially aligned information from adjacent scales. EviRegionSearch performs \textbf{question-conditioned exploration} over the ROI identified by CrossMagLoc, integrates morphological observations from the selected ROI with the accumulated evidence, identifies the remaining evidence requirement, and adaptively determines the next navigation action ({Move}, {Zoom In}, {Zoom Out}, or {Stop}) to select the next region.
}\label{fig2}
\end{figure*}

\textbf{PathPilot} is an iterative question-conditioned WSI navigation agent composed of two complementary and sequentially coupled components (Fig.~\ref{fig2}):
(1) \textbf{Cross-Magnification Localization Tool (CrossMagLoc)}, which performs question-agnostic localization within the current ROI and identifies the diagnostically most relevant region (see Sec.~\ref{sec:CrossMagLoc}).
(2) \textbf{ Evidence-Adaptive Region Search Tool (EviRegionSearch)}, which performs question-conditioned exploration over these candidates by progressively accumulating relevant evidence and determining where and how to continue according to the remaining evidence requirement (see Sec.~\ref{sec:EviRegionSearch}).
\section{Method}

\subsection{Overall Workflow.}

As shown in Fig. \ref{fig2} and Eq. \ref{eq:pathpilot_workflow}, given a multi-magnification WSI and a pathology question $Q$, PathPilot starts from the thumbnail $I^0$ and iteratively performs question-conditioned navigation across ROIs and magnifications. At each navigation step $t$, CrossMagLoc generates a question-agnostic heatmap $H^t$ from $I^{t-1}$, highlighting regions important for disease diagnosis. EviRegionSearch then uses this heatmap and the action selected from the last step to obtain the ROI $I^t$ for its examination. It then incorporates the observation from this ROI $I^t$ to update the accumulated question-relevant evidence $\mathcal{E}^t$ and remaining evidence requirement $\mathcal{R}^t$, and determines the next navigation action (Move, Zoom In, Zoom Out, or Stop). For a non-stop action, the navigation state is updated, and the selected action is executed at the next iteration to continue evidence accumulation. When sufficient evidence has been accumulated, EviRegionSearch selects Stop and the accumulated evidence is used to generate the final answer.

\begin{equation}
\resizebox{1.0\linewidth}{!}{
$
\begin{aligned}
I^{0}
&\xrightarrow{\mathrm{CrossMagLoc}}
H^{1}
\xrightarrow[\;Q\;]{\mathrm{EviRegionSearch}}
\\[-0.2ex]
&\begin{cases}
\big(I^{1}, \mathcal{E}^{1}, \mathcal{R}^{1}\big)
\xrightarrow{\mathrm{CrossMagLoc}}
H^{2}
\xrightarrow[\;Q\;]{\mathrm{EviRegionSearch}}
\cdots
\rightarrow
\big(I^{t}, \mathcal{E}^{t}, \mathcal{R}^{t}\big),
& \text{case (1)},
\\[1ex]
\mathrm{Answer}\big(Q\big),
& \text{case (2)}.
\end{cases}
\\[1.2ex]
&\text{case (1): the accumulated evidence is insufficient,}
\\[0.5ex]
&\text{case (2): the accumulated evidence is sufficient.}
\end{aligned}
$
}
\label{eq:pathpilot_workflow}
\end{equation}

\label{eq:mmnav_workflow}

\subsection{Cross-Magnification Localization Tool (CrossMagLoc).}
\label{sec:CrossMagLoc}

Learning question-agnostic diagnostic localization from pathologist viewing behavior faces two challenges:  (1) \textbf{navigation supervision is isolated} across magnifications, making per-magnification heatmaps much sparser and harder to learn than heatmaps aggregated across magnifications; and (2) \textbf{annotations are incomplete}, because pathologists do not need to inspect all diagnostically relevant regions to reach a diagnosis, especially when multiple regions provide similar diagnostic information.

To address the first challenge, CrossMagLoc introduces learnable magnification tokens to capture the diagnostic focus at each magnification. These tokens are used in two complementary blocks: an \textbf{Intra-Magnification Cross-Attention Block}, which uses the current magnification token to incorporate magnification-specific information into the feature representation, and a \textbf{ Cross-Magnification Context Fusion Block}, which uses adjacent magnification tokens to integrate spatially corresponding contextual information from adjacent magnification levels.

To address the second challenge, we further introduce a \textbf{Navigation-Driven Supervision Loss (NDSL)} to improve learning from sparse and potentially incomplete navigation annotations.

In the following, we first use a representative case to illustrate CrossMagLoc, which consists of four components:
(1) a shared WSI Encoder,
(2) an Intra-Magnification Cross-Attention Block,
(3) a Cross-Magnification Context Fusion Block, and
(4) a Localization Decoder. 
Then we introduce the NDSL used to supervise CrossMagLoc.

\subsubsection{Shared WSI Encoder.}

Given an ROI $I^m$ at magnification $m$, a shared frozen pathology encoder ${F}_{\mathrm{enc}}$ encodes the image into patch tokens,
\begin{equation}
\mathbf{X}^{m}
=
{F}_{\mathrm{enc}}({I}^{m})
=
[\mathbf{x}^{m}_{1},\mathbf{x}^{m}_{2},\ldots,\mathbf{x}^{m}_{N}]
\in\mathbb{R}^{N\times D},
\end{equation}
where $N$ denotes the number of patch tokens and $D$ the feature
dimension. Each token $\mathbf{x}^{m}_{i}$ corresponds to a fixed
non-overlapping spatial region in the original image ${I}^{m}$.
${F}_{\mathrm{enc}}$ is shared across all magnification levels and remains frozen during CrossMagLoc training.

\subsubsection{Intra-Magnification Cross-Attention Block (IMCA).}

The \textbf{Intra-Magnification Cross-Attention Block (IMCA)} aims to explicitly encode the magnification information into the representation of each individual magnification. For each magnification level $m$, we introduce a learnable magnification-specific token $\mathbf{t}^{m}$ and use it to interact with the patch representation $\mathbf{X}^{m}$ through cross-attention. In this way, the representation learned at each scale is explicitly aware of its corresponding magnification rather than being treated as generic visual features.

Specifically, the magnification token $\mathbf{t}^{m}$ serves as the query, while the patch tokens $\mathbf{X}^{m}$ serve as the keys and values. The resulting magnification-aware information is then reinjected into $\mathbf{X}^{m}$ through a gated residual connection:
\begin{equation}
\small
\mathbf{X}^m \leftarrow \mathbf{X}^m + \gamma \cdot \text{Attn}(\mathbf{t}^m, \mathbf{X}^m, \mathbf{X}^m),
\end{equation}
where $\gamma$ is a learnable scalar controlling the contribution of the magnification-aware information.

\subsubsection{Cross-Magnification Context Fusion Block (CMCF).}

While IMCA models magnification information \textbf{within each individual scale}, the \textbf{Cross-Magnification Context Fusion Block (CMCF)} further introduces complementary information from adjacent magnifications. This is particularly important because navigation supervision is isolated across magnifications, making the localization signal at each individual scale relatively sparse. CMCF therefore allows the representation at magnification $m$ to incorporate spatially corresponding information from its neighboring magnifications $m-1$ and $m+1$.

Specifically, for each magnification $m$, we first spatially align the representations from its adjacent magnifications to the current scale, obtaining $\hat{\mathbf{X}}^{m-1}$ from the lower magnification and $\hat{\mathbf{X}}^{m+1}$ from the higher magnification through spatial interpolation. Their magnification-specific information is then extracted through cross-attention using the corresponding tokens $\mathbf{t}^{m-1}$ and $\mathbf{t}^{m+1}$. The resulting adjacent-magnification features are fused into the current representation $\mathbf{X}^{m}$ through a weighted residual connection:
\begin{equation}
\resizebox{0.85\linewidth}{!}{
$ \mathbf{X}^m \leftarrow \mathbf{X}^m + u \cdot \text{Attn}(\mathbf{t}^{m-1}, \hat{\mathbf{X}}^{m-1}, \hat{\mathbf{X}}^{m-1}) + w \cdot \text{Attn}(\mathbf{t}^{m+1}, \hat{\mathbf{X}}^{m+1}, \hat{\mathbf{X}}^{m+1}), $
}
\label{eq:cross_mag_fuse}
\end{equation}
where $u$ and $w$ are learnable weights controlling the contributions from the lower- and higher-magnification contexts, respectively.

\subsubsection{Localization Decoder.}

Finally, the updated representation $\mathbf{X}^{m}$ is fed into a
lightweight shared CNN decoder ${D}_{\mathrm{cnn}}(\cdot)$ to
generate the magnification-specific localization heatmap
\begin{equation}
H^{m} = {D}_{\mathrm{cnn}}(\mathbf{X}^{m}),
\end{equation}
where $H^m\in\mathbb{R}^{H\times W\times 1}$ represents
question-agnostic diagnostic relevance within the input ROI
at magnification $m$.

\subsubsection{Navigation-Driven Supervision Loss (NDSL).} 
A key limitation of navigation annotations is that the ground truth $G^m$ is typically sparse and subjective  \cite{nan2025deep}. To mitigate this, we propose a compact NDSL, which integrates a soft Dice loss~ \cite{milletari2016v} and a soft focal loss~ \cite{lin2017focal}. This formulation encourages alignment with existing annotations while avoiding excessive penalties on unlabeled but clinically diagnostic regions. Combined with a weighted $\ell_{1}$ loss $\mathcal{L}_{1}$ as the primary supervision, the NDSL objective is mathematically formulated as:
\begin{equation}
\small
\mathcal{L}_{\mathrm{nav}}
= \lambda_{l}\,\mathcal{L}_{1}(H^m,G^m)+\lambda_{\mathrm{dice}}\,\mathcal{L}_{\mathrm{dice}}(H^m,G^m)
+ \lambda_{\mathrm{sig}}\,\mathcal{L}_{\mathrm{sig}}(H^m,G^m),
\end{equation}

\subsection{Evidence-Adaptive Region Search Tool (EviRegionSearch).}
\label{sec:EviRegionSearch}

While CrossMagLoc generates question-agnostic heatmaps highlighting regions important for disease diagnosis, EviRegionSearch performs question-conditioned exploration to determine what additional visual evidence is required and how the navigation should proceed.

Starting from the thumbnail image $I^0$, EviRegionSearch uses the heatmap generated by CrossMagLoc to select the most diagnostically relevant ROI for initial examination. It then accumulates question-relevant evidence from this ROI and determines the first navigation action. Each subsequent step begins by using the corresponding heatmap to execute the previous step's action and obtain the next ROI for examination. Each ROI is retrieved directly from the original WSI at the target magnification, preserving the native image resolution.

Let $I^{t,m_t}$ denote the ROI examined at step $t$ under magnification $m_t$. We denote the accumulated question-relevant evidence up to step $t$ as $\mathcal{E}^{t}$ and the remaining evidence requirement after this step as $\mathcal{R}^{t}$. EviRegionSearch updates $\mathcal{E}^{t}$ and $\mathcal{R}^{t}$ based on the observation from $I^{t,m_t}$, the question $Q$, and the previous states $\mathcal{E}^{t-1}$ and $\mathcal{R}^{t-1}$, and then determines the next navigation action accordingly.

As shown in Fig.~\ref{fig2}, once the ROI has been obtained, EviRegionSearch proceeds through two sequential steps: \textbf{Evidence Requirement Reasoning}, which updates the accumulated evidence and determines what visual evidence remains required, and \textbf{Evidence-to-Action Translation}, which translates the remaining evidence requirement into the next navigation action. The overall process is summarized as:

\begin{equation}
\small
\left(
Q,\,
\mathcal{E}^{t-1},
\mathcal{R}^{t-1},
I^{t,m_t}
\right)
\xrightarrow[\mathrm{Reasoning}]
{\mathrm{Evidence~Requirement}}
\left(
\mathcal{E}^{t},
\mathcal{R}^{t}
\right)
\xrightarrow[\mathrm{Translation}]
{\mathrm{Evidence\text{-}to\text{-}Action}}
Op^{t}.
\label{eq:ques_region_search}
\end{equation}
Here, $\mathcal{E}^{t-1}$ and $\mathcal{R}^{t-1}$ denote the accumulated evidence and remaining evidence requirement from the previous step, respectively, and $Op^t$ denotes the navigation action determined at the current step for execution at the next step, unless Stop is selected.

\textbf{Evidence Requirement Reasoning.}
The Evidence Requirement Reasoning step analyzes the current ROI together with the accumulated question-relevant evidence to determine:

\begin{itemize}
    \item \textit{What evidence has been obtained.}
    \item \textit{What visual information is still required.}
\end{itemize}

To obtain question-relevant observations without directly exposing the VLM to answer-oriented wording, we first transform the original question $Q$ into compact visual guidance $\tilde{Q}$ describing the relevant visual scale, context, and morphology to inspect (prompt design is provided in the Supplementary Material). Given the current ROI ${I}^{t,m_t}$, the VLM generates a question-focused morphological observation:

\begin{equation}
\small
D^{t,m_t}
=
\mathrm{VLM}
\left(
{I}^{t,m_t},
\tilde{Q}
\right).
\label{eq:vlm_description}
\end{equation}

Here, $D^{t,m_t}$ denotes the question-focused morphological observation of the current ROI. The morphological observation is then jointly considered with the accumulated question-relevant evidence $\mathcal{E}^{t-1}$ and the remaining evidence requirement $\mathcal{R}^{t-1}$ under the current question $Q$ to update $\mathcal{E}^{t}$ and $\mathcal{R}^{t}$:

\begin{equation}
\small
\mathcal{E}^{t},
\mathcal{R}^{t}
=
\mathrm{LLM}^{\mathrm{Reason}}
\left(
Q,D^{t,m_t},
\mathcal{E}^{t-1},
\mathcal{R}^{t-1}
\right).
\label{eq:evidence_accumulation}
\end{equation}

\textbf{Evidence-to-Action Translation.}
Based on the updated accumulated question-relevant evidence $\mathcal{E}^{t}$ and remaining evidence requirement $\mathcal{R}^{t}$, the Evidence-to-Action Translation step determines the next navigation action:

\begin{equation}
\small
Op^{t}
=
\mathrm{LLM}^{\mathrm{Select}}
\left(
Q,
\mathcal{E}^{t},
\mathcal{R}^{t}
\right).
\label{eq:evidence_reasoning}
\end{equation}

Here, $Op^{t}$ is selected from \textbf{Move}, \textbf{Zoom In}, \textbf{Zoom Out}, and \textbf{Stop}. For each non-stop action, the selected action determines the next navigation state and how the next ROI is obtained for examination.

\textbf{(i) Move.}
When additional evidence is required from a different location,
EviRegionSearch reuses the current heatmap and selects the
highest-scoring unexamined region for examination at the
current magnification.

\textbf{(ii) Zoom In.}
When finer morphological details are required, CrossMagLoc
generates a heatmap for the current ROI at the next step.
EviRegionSearch uses this heatmap to select the most
diagnostically relevant region and retrieves it at a higher
magnification for examination.

\textbf{(iii) Zoom Out.}
When broader tissue context is required, EviRegionSearch
retrieves the corresponding lower-magnification view of the
current ROI directly from the original WSI for examination
at the next step.

\textbf{(iv) Stop.}
The agent terminates navigation when the accumulated
question-relevant evidence $\mathcal{E}^{t}$ is sufficient
for answering $Q$, corresponding to
$\mathcal{R}^{t}=\varnothing$ and $Op^{t}=Op_{\mathrm{stop}}$.

\begin{figure*}[th]
  \centering
\includegraphics[width=0.95\textwidth]{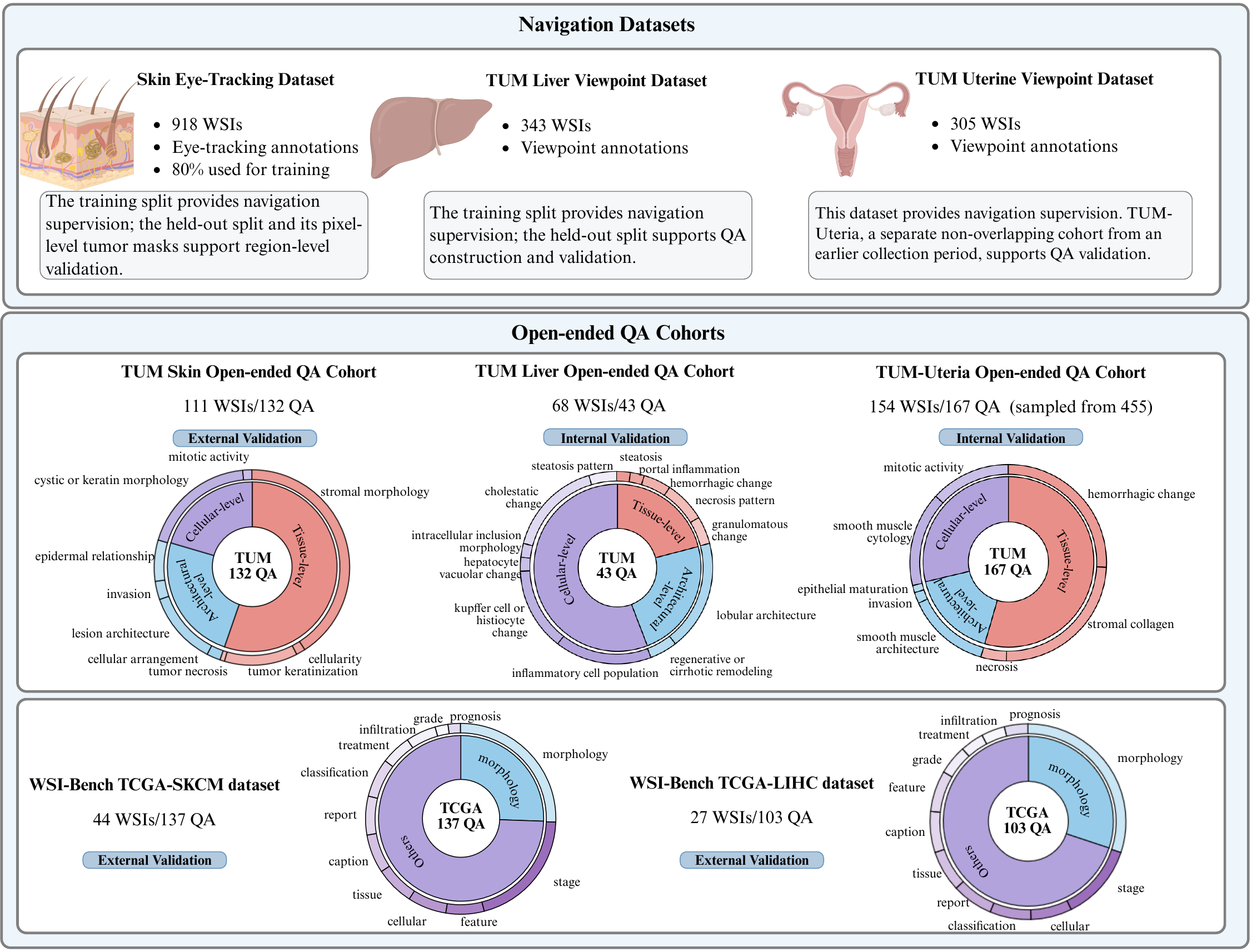}
\caption{
\textbf{Overview of the datasets and evaluation cohorts used in this study.}
\textbf{Top:} Three navigation datasets provide pathologist viewing supervision for CrossMagLoc, including the Skin Eye-Tracking dataset and the TUM Liver and TUM Uterine Viewpoint datasets. The Skin dataset additionally provides pixel-level tumor annotations for region-level localization evaluation.
\textbf{Bottom:} Five open-ended QA cohorts are used for end-to-end PathPilot evaluation, comprising the internal TUM Liver and TUM-Uteria cohorts and the external TUM Skin, WSI-Bench TCGA-SKCM, and WSI-Bench TCGA-LIHC cohorts. For the TUM cohorts, the circular plots summarize the predefined H\&E-visible findings grouped into tissue-level, architectural, and cellular categories; for the WSI-Bench cohorts, they summarize the distributions of question categories.
} \label{fig:dataset}
\end{figure*}

\section{Experiments}
\label{sec:experiments}

We quantitatively evaluate PathPilot from three complementary perspectives: the \textbf{overall WSI analysis performance} of the complete framework, the \textbf{navigation performance} of CrossMagLoc, and the \textbf{evidence-adaptive exploration performance} of EviRegionSearch.

We first evaluate the overall WSI analysis performance of PathPilot through open-ended QA \textbf{on five cohorts}, including \textbf{two internal cohorts} (TUM Liver and TUM-Uteria~\cite{li2026paired}) and \textbf{three external cohorts} (TUM Skin, WSI-Bench TCGA-SKCM~\cite{liang2025wsi}, and WSI-Bench TCGA-LIHC~\cite{liang2025wsi}). For the TUM cohorts, performance is further analyzed across tissue-level, architectural, and cellular findings with different visual evidence requirements.

We then evaluate the two navigation components separately. CrossMagLoc is evaluated through region-level localization performance, including alignment with pathologist region prioritization and annotated tumor regions across magnifications. EviRegionSearch is evaluated through controlled studies of evidence requirement reasoning and evidence-to-action translation to examine its evidence-adaptive exploration across regions and magnifications.

Beyond the quantitative evaluation, we provide two complementary qualitative analyses: a \textbf{pathologist-in-the-loop assessment of CrossMagLoc} and a \textbf{representative PathPilot navigation workflow} illustrating how question-conditioned navigation progressively acquires evidence across regions and magnifications. Finally, additional ablation studies examine the loss and architectural components of CrossMagLoc, together with the visual guidance and model configurations of EviRegionSearch.

The remainder of this section is organized as follows. The datasets and evaluation settings are introduced in Sec.~\ref{sec:datasets} and \ref{sec:exp_setting}, followed by the end-to-end QA evaluation in Sec.~\ref{sec:qa_exp} and the evaluation of CrossMagLoc in Sec.~\ref{sec:nav_exp} and EviRegionSearch in Sec.~\ref{sec:eviregionsearch_eval}. Qualitative analyses and ablation studies are presented in Sec.~\ref{sec:qualitative} and Sec.~\ref{sec:ablation}, respectively.

\subsection{Datasets.}
\label{sec:datasets}
As summarized in Fig.~\ref{fig:dataset}, the datasets are organized into two categories according to their roles: \textbf{navigation datasets and QA cohorts}. The navigation datasets, including the \hyperref[subsec:skin-eye-tracking]{Skin Eye-Tracking dataset}, the \hyperref[subsec:liver-screen-tracking]{TUM Liver Viewpoint dataset}, and the \hyperref[subsec:uterine-screen-tracking]{TUM Uterine Viewpoint dataset}, are used to train and validate CrossMagLoc. The QA cohorts, including the  \hyperref[subsec:tum-liver-qa]{TUM Liver QA cohort}, \hyperref[subsec:tum-uteria-qa]{TUM-Uteria QA cohort}, \hyperref[subsec:tum-skin-qa]{TUM Skin QA cohort}, \hyperref[subsec:tcga-skcm-qa]{WSI-Bench TCGA-SKCM}, and \hyperref[subsec:tcga-lihc-qa]{WSI-Bench TCGA-LIHC} are used to evaluate the complete PathPilot framework through end-to-end open-ended question answering. The TUM navigation datasets and QA cohorts are private, except for the publicly available TUM-Uteria. The two dataset categories are described in detail below.

\subsubsection{Navigation Datasets.}
Because the available annotation formats differ across cohorts, the three navigation datasets serve different roles in our experiments. The Skin Eye-Tracking dataset is used for CrossMagLoc training and, owing to its additional pixel-level tumor annotations, its held-out split is further used for direct region-level navigation evaluation. The TUM Liver Viewpoint dataset is used for CrossMagLoc training, while its held-out cases are used to construct the TUM Liver QA cohort for end-to-end PathPilot evaluation rather than for region-level navigation benchmarking. The TUM Uterine Viewpoint dataset is used only for CrossMagLoc training. For uterine QA evaluation, we instead use the publicly available TUM-Uteria cohort from the same institution, providing an openly accessible benchmark for evaluating the complete PathPilot framework and increasing confidence in the internal QA validation.

\phantomsection
\label{subsec:skin-eye-tracking}
\textbf{Skin Eye-Tracking dataset}  \cite{nan2025deep} contains 918 WSIs, including 246 nevus, 517 basal cell carcinoma (BCC), 105 melanoma, and 50 squamous cell carcinoma (SCC) cases. Pathologists' gaze locations and viewing magnifications are recorded during slide examination and converted into multi-magnification navigation heatmaps for CrossMagLoc supervision. \textbf{Pixel-level tumor annotations} are additionally available. We adopt a case-level 80\%:20\% training--test split: the training split is used to train CrossMagLoc, while the held-out test split is reserved for direct region-level validation through pathologist-attention alignment and tumor-region overlap.

\phantomsection
\label{subsec:liver-screen-tracking}
\textbf{TUM Liver Viewpoint dataset} contains 343 WSIs \textbf{collected during routine pathological diagnosis} at our hospital. It therefore reflects the clinical distribution of liver findings encountered in daily diagnostic practice rather than a deliberately constructed transfer distribution. During diagnosis, the visible regions, viewing duration, and magnification used by pathologists are recorded and converted into multi-magnification navigation heatmaps for CrossMagLoc supervision. We adopt a case-level 80\%:20\% split for CrossMagLoc training, with the held-out portion additionally used to construct the TUM Liver QA cohort described below.

\phantomsection
\label{subsec:uterine-screen-tracking}
\textbf{TUM Uterine Viewpoint dataset} is likewise \textbf{collected during routine pathological diagnosis} at our hospital and reflects the clinical distribution of uterine cases encountered in daily practice. Viewpoint records are processed using the same procedure as for the TUM Liver Viewpoint dataset and provide navigation supervision for CrossMagLoc. This dataset is temporally distinct from TUM-Uteria, which was collected before viewpoint was introduced into the routine diagnostic workflow and therefore contains no navigation annotations.

\subsubsection{Open-ended QA Cohorts.}

The five open-ended QA cohorts, including two internal cohorts and three external cohorts, are used \textbf{only for end-to-end evaluation}, without any task-specific training or fine-tuning. This avoids adapting PathPilot to cohort-specific question formulations, answer distributions, or benchmark priors, ensuring that the evaluation primarily reflects its ability to acquire and integrate relevant visual evidence from WSIs for previously unseen questions.

Moreover, unlike option-based QA (or closed-set QA)  \cite{lu2024multimodal, chen2024wsi} with predefined answer options, open-ended QA requires \textbf{free-form pathology answer generation}, thereby providing a more stringent evaluation of whether PathPilot can independently identify relevant visual evidence and translate it into diagnostically meaningful responses without relying on cues from candidate answers. Internal validation is conducted on TUM Liver and TUM-Uteria, whereas external validation is conducted on TUM Skin, WSI-Bench TCGA-SKCM, and WSI-Bench TCGA-LIHC.

\phantomsection
\label{subsec:tum-liver-qa}
\textbf{TUM Liver Open-ended QA cohort} is constructed from the held-out portion of the TUM Liver Viewpoint dataset and contains 43 open-ended QA pairs from 68 WSIs. The questions focus on visually assessable H\&E findings in non-tumor/background liver pathology, including steatosis, portal inflammation, necrosis, granulomatous changes, architectural alterations, and cellular morphology.

\phantomsection
\label{subsec:tum-uteria-qa}
\textbf{TUM-Uteria Open-ended QA cohort}  \cite{li2026paired} is collected at the same institution but during an earlier period before viewpoint was introduced. It therefore contains no navigation annotations and is used only for QA validation. The full dataset contains 455 WSIs paired with pathology reports, from which 154 WSIs are used to construct 167 open-ended QA pairs covering diagnostically relevant H\&E-visible findings in uterine pathology.

\phantomsection
\label{subsec:tum-skin-qa}
\textbf{TUM Skin Open-ended QA cohort} contains 111 WSIs collected from routine clinical pathology practice at our hospital. From the corresponding pathology reports, we construct 132 open-ended QA pairs targeting predefined H\&E-visible findings. Questions relying on non-H\&E information, uncertain findings, or unsupported inference are excluded, and all retained QA pairs are manually reviewed and independently verified by pathologists.

\phantomsection
\label{subsec:tcga-skcm-qa}
\textbf{WSI-Bench TCGA-SKCM cohort}  \cite{liang2025wsi} contains 137 open-ended questions across 44 WSIs in the official test set. We use its morphology subset as the primary evaluation target, since these questions most directly depend on retrieving and integrating H\&E-visible evidence from the WSI and therefore provide a focused assessment of visual evidence acquisition. The morphology subset contains 35 questions across 9 WSIs. The remaining WSI-Bench task categories are relatively sparsely represented in this cohort-specific test set and include tasks for which varying amounts of answer-relevant diagnostic, clinical, or immunohistochemical information are explicitly provided in the question itself. We therefore report performance over the complete question set as supporting evidence of robustness to a broader question distribution, while morphology-focused performance serves as the primary visually grounded evaluation.

\phantomsection
\label{subsec:tcga-lihc-qa}
\textbf{WSI-Bench TCGA-LIHC cohort}  \cite{liang2025wsi} contains 103 open-ended questions across 27 WSIs in the official test set. Following the same evaluation setting as TCGA-SKCM, morphology questions are used as the primary visually grounded evaluation, comprising 31 questions across 7 WSIs, while performance over the complete question set is reported as supporting evidence.

For the TUM QA cohorts, questions are grouped according to their predominant visual evidence requirements into tissue-level, architectural, and cellular categories. These categories \textbf{describe different clinically relevant aspects} of pathological assessment rather than fixed magnification levels, as individual questions may require evidence from multiple spatial regions and magnifications. Their distributions and detailed question settings are summarized in Fig.~\ref{fig:dataset} and Table~\ref{table:qa_setting}.

\subsection{Experimental Settings.}
\label{sec:exp_setting}

\subsubsection{Data Preparation.}
To make the corresponding data suitable for model training and evaluation, different preparation procedures are required for the navigation and QA cohorts: For navigation, the raw pathologist viewing records consist of sequential gaze or viewport observations across different magnifications and therefore need to be spatially aligned and aggregated into magnification-specific supervision for CrossMagLoc. For QA evaluation, the TUM cohorts provide pathology reports rather than predefined question--answer pairs, requiring report-grounded QA construction for evaluating the complete PathPilot framework.

\textbf{Navigation data processing.}
We construct multi-magnification navigation supervision from \textbf{eye-tracking and viewpoint} records. Eye-tracking records specify gaze locations and magnifications, while viewpoint records specify visible rectangular WSI regions, viewing durations, and magnifications. Observations are projected onto WSI coordinates, weighted equally for eye-tracking and by viewing duration for viewpoint, spatially smoothed, and normalized to $[0,1]$.
For thumbnail supervision, all valid observations are aggregated into a slide-level heatmap. At other magnifications, heatmaps are constructed in magnification-specific coordinates and paired with $256\times256$ image patches. Each spatially matched heatmap region is extracted and independently normalized; patches are retained only if more than 30\% of pixels have normalized values above 0.5. Analysis of all navigation records shows that diagnostic viewing occurs almost exclusively between 1.25$\times$ and 10$\times$. Accordingly, we construct navigation supervision at the thumbnail, 1.25$\times$, 2.5$\times$, 5$\times$, and 10$\times$ levels for training CrossMagLoc.

\textbf{QA data construction.}
For the TUM QA cohorts, pathologists define organ-specific banks of open-ended questions addressing morphological findings directly assessable on H\&E-stained WSIs. Given each candidate question and its corresponding pathology report, Qwen3-14B identifies explicit supporting report evidence. Questions are excluded if supporting evidence is insufficient, if they depend on immunohistochemistry, molecular findings, staging information, clinical history, or uncertain findings, or if they require unsupported diagnostic inference. Reference answers are generated exclusively from the identified evidence. Automatic validation requires this evidence to match an exact contiguous span of the source report and remain semantically consistent with the answer; overly broad evidence is reduced to the shortest span containing the required morphological object and feature. The resulting QA pairs undergo manual review and independent verification by pathologists. Their distributions and detailed question settings are provided in Fig.~\ref{fig:dataset} and Table~\ref{table:qa_setting}.

Clinical pathological assessment integrates tissue context, structural organization, and cellular morphology across spatial scales \cite{schmitz2021multi,pati2022hierarchical}. To reflect these \textbf{complementary clinical assessment requirements}, we group TUM questions into \textbf{tissue-level, architectural-level, and cellular-level} categories according to their predominant visual evidence requirements. These categories concern findings distributed across broader tissue regions, the organization and spatial relationships of pathological structures, and localized fine-grained cellular morphology, respectively. They indicate predominant evidence requirements rather than fixed magnifications, as individual questions may require evidence across multiple regions and magnification levels. For the public WSI-Bench TCGA-SKCM and TCGA-LIHC cohorts, the official question--answer pairs are used directly without additional QA construction.

\subsubsection{Training Details.}
For CrossMagLoc, we use our cellular-scale pathology foundation model Hireca~ \cite{Liu2025TowardsCI} as the feature extractor and a U-Net~ \cite{ronneberger2015u} decoder without skip connections as $\mathcal{D}_{\mathrm{cnn}}$. The decoder progressively upsamples the feature map through four $2\times$ upsampling blocks, each consisting of a transposed convolution followed by a $3\times3$ convolution, InstanceNorm, and ReLU, and ends with a $1\times1$ convolution to generate the single-channel navigation heatmap. We set $\lambda_{\mathrm{l}}=0.1$ with a foreground weight of 2.0 and $\lambda_{\mathrm{dice}}=\lambda_{\mathrm{sig}}=1$. Both the input and navigation heatmaps have a resolution of $256\times256$. Training uses Adam with a batch size of 64, a learning rate of $10^{-4}$, and a weight decay of 0.01, for 1,000 training iterations. To avoid magnification imbalance, each batch contains equal numbers of patches from each magnification, with the per-scale sample count determined by the smallest available magnification-specific set. Based on the eye- and viewpoint statistics, navigation heatmaps are generated at the thumbnail, 1.25$\times$, 2.5$\times$, 5$\times$, and 10$\times$ levels. A case-level 80\%:20\% training-test split is used.

For EviRegionSearch, we use Qwen3-14B~\cite{yang2025qwen3} as the LLM and Patho-R1-7B~\cite{zhang2026patho} as the VLM. Qwen3-14B uses deterministic decoding with a temperature of 0 and maximum generation lengths of 1,536 tokens for navigation planning and 2,536 tokens for final answering. Patho-R1-7B is loaded locally with automatic precision and device allocation and uses greedy decoding with up to 1,024 new tokens. During initialization, we retrieve and examine six low-magnification candidate ROIs using the thumbnail-level navigation heatmap produced by CrossMagLoc. These candidates and their observations are retained for subsequent navigation, eliminating the need to access the thumbnail-level heatmap again. Among these candidates, the ROI with the highest mean navigation score serves as the starting point for subsequent exploration, where the mean navigation score is the average heatmap value within the corresponding patch. After initialization, each navigation step retrieves and examines only one ROI, with the LLM determining the next navigation action based on the remaining evidence requirements. Each newly retrieved $256\times256$ ROI is described by the VLM only once, with duplicate image paths removed before VLM inference. Each question is limited to 30 navigation tool calls. All ROI observations are retained in the Memory Bank, while only the 20 most recent ROI descriptions are included in the LLM context for each navigation decision.

All experiments are conducted on two NVIDIA H100 80GB HBM3 GPUs, with one GPU reserved for LLM inference when required. Our implementation uses PyTorch 2.9.0, whereas baseline methods use the PyTorch versions specified by their official implementations.

\subsubsection{Open-ended QA Evaluation Protocol.} 
Open-ended pathology QA cannot be evaluated as directly as option-based (or closed-ended) choice QA, where predictions can be matched against predefined answer options. Its free-form responses can express the same clinically correct finding with substantially different wording. To address this, prior pathology QA studies have adopted expert- or LLM-based semantic assessment for free-form responses~ \cite{seyfioglu2024quilt,liang2025wsi,lu2024multimodal}. Following this practice, we use a local Qwen-based pathology QA judge and further refine the grading granularity for our report-grounded QA setting, enabling \textbf{a finer-grained assessment of semantic correctness} beyond binary correctness. Specifically, our evaluation consists of two components: a Qwen-based pathology QA grading procedure that assigns graded semantic correctness scores to individual responses, and a set of QA performance metrics that aggregate these scores to quantitatively measure PathPilot's WSI understanding capability. We further assess the evaluation protocol using an independent LLM, as detailed in Sec.~\ref{sec:judge_validation}.

\textbf{(i) Qwen-based pathology QA grading.} 
Following prior graded semantic evaluation~ \cite{seyfioglu2024quilt,liang2025wsi}, the judge compares each agent answer with the question, report-derived reference answer, and corresponding report evidence. Responses are assigned one of five scores: 1.0 (fully correct), 0.7 (mostly correct), 0.5 (partially correct), 0.3 (weakly related), or 0.0 (incorrect). The reference answer is treated as a semantic exemplar rather than an exact lexical target, allowing clinically valid paraphrases and concise correct answers while penalizing incompatible findings, incorrect polarity, hallucinated content, and irrelevant information. To improve grading consistency, the evaluator follows an audit--grading--verification procedure that checks target coverage, substantive agreement, irrelevant content, and hallucinated findings before assigning and verifying the final score. 

\textbf{(ii) QA performance metrics.} 
Based on the semantic scores, we report three complementary metrics. \textbf{Graded Accuracy (GACC)}, the primary metric, is the mean semantic correctness score across all questions. \textbf{Passable-or-Better Rate (PBR)} is the proportion of answers scoring at least 0.7, corresponding to mostly or fully correct responses, whereas \textbf{Incorrect Answer Rate (IAR)} is the proportion scoring 0.0, corresponding to completely incorrect or irrelevant responses. PBR complements GACC by capturing substantively correct answers without requiring close lexical agreement with a single reference formulation.

\begin{figure*}[th]
  \centering
\includegraphics[width=1.0\textwidth]{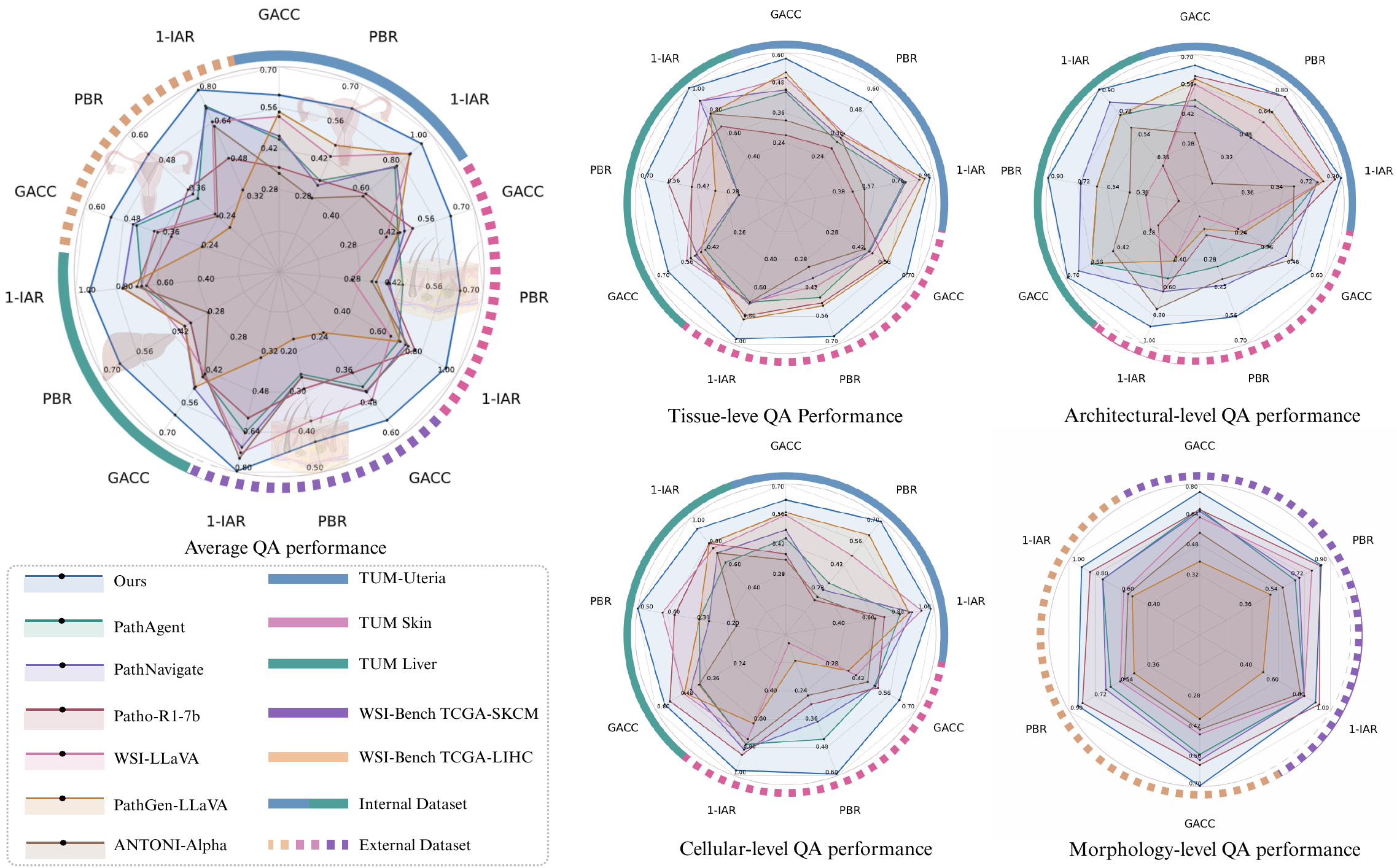}
\caption{ \textbf{QA performance of PathPilot across internal and external validation cohorts.} Comparisons include pathology-specialized thumbnail-based VLMs (PathGen-LLaVA and Patho-R1-7B), patch-based WSI VLMs (WSI-LLaVA and ANTONI-Alpha), and pathology agents (PathAgent and PathNavigate). The left radar plot summarizes average performance across all questions on two internal cohorts (TUM-Uteria and TUM Liver) and three external cohorts (TUM Skin, WSI-Bench TCGA-SKCM, and WSI-Bench TCGA-LIHC). The four right-hand plots further assess tissue-level, architectural-level, and cellular-level QA on the three TUM cohorts, and morphology-focused QA on the two WSI-Bench cohorts, distinguishing overall QA performance from performance on specific pathological assessment tasks. Each cohort is evaluated using GACC, PBR, and $1-\mathrm{IAR}$, with higher values indicating better performance on all axes. Colored outer arcs identify cohorts; solid and dashed arcs denote internal and external validation, respectively. The plots summarize the quantitative results in Tables~\ref{results:tum-uteria}--\ref{results:tcga_external}. }
\label{fig:performance}
\end{figure*}

\newsavebox{\tablebox}
\begin{table*}[t]
\centering
\setlength{\fboxsep}{0pt}

\caption{Detailed QA settings for the three TUM datasets, including the three predefined question types (tissue-level, architectural-level, and cellular-level) and the specific question labels and corresponding questions within each type.} \label{table:qa_setting}
\begin{lrbox}{\tablebox}
\begin{tabular}{llll}
\toprule[1.5pt]
\toprule[1.5pt]
\textbf{Dataset}
& \textbf{Question label}
& \textbf{Question type}
& \textbf{Question}\\
\toprule[1pt]
\multirow{10}{*}{\textbf{TUM Skin}}&stromal morphology&Tissue-level&What stromal change is observed in the subcutaneous tissue?\\
&cellularity&Tissue-level&Which compartment exhibits reduced cellularity within the fibrotic lesion?\\
&tumor keratinization&Tissue-level&What keratinization pattern is present in the tumor?\\
&tumor necrosis&Tissue-level&What necrotic change is present in the tumor?\\
&cellular arrangement&Architectural-level&How are the lesional melanocytes arranged?\\
&lesion architecture&Architectural-level&How is the epithelial structure organized around the fistula-like invagination?\\
&invasion&Architectural-level&Which tissue or compartment is invaded by the tumor?\\
&epidermal relationship&Architectural-level&What relationship is present between the tumor and the epidermis?\\
&cystic or keratin morphology&Cellular-level&How is the epidermal cystic structure characterized?\\
&mitotic activity&Cellular-level&How is mitotic activity described in the tumor?\\
\toprule[1pt]
\multirow{13}{*}{\textbf{TUM Liver}}&steatosis&Tissue-level&What is the broad degree of steatosis?\\
&portal inflammation&Tissue-level&How is the inflammatory infiltrate distributed within the portal tracts?\\
&hemorrhagic change&Tissue-level&How is hemorrhage distributed within the liver tissue?\\
&necrosis pattern&Tissue-level&How are the necrotic areas distributed within the liver parenchyma?\\
&granulomatous change&Tissue-level&What cellular components form the granulomatous reaction?\\
&lobular architecture&Architectural-level&How is the hepatic lobular architecture organized?\\
&regenerative or cirrhotic remodeling&Architectural-level&How is the adjacent liver parenchyma remodeled in the context of cirrhosis?\\
&inflammatory cell population&Cellular-level&Which inflammatory cell populations compose the infiltrate within the cyst wall?\\
&kupffer cell or histiocyte change&Cellular-level&How are the hepatic macrophage-lineage cells morphologically altered?\\
&hepatocyte vacuolar change&Cellular-level&What cytoplasmic vacuolar alteration is present in the hepatocytes?\\
&intracellular inclusion morphology&Cellular-level&What intracellular inclusion is identified within the hepatocytes?\\
&cholestatic change&Cellular-level&How is cholestatic change manifested in the liver tissue?\\
&steatosis pattern&Cellular-level&What pattern of steatosis is seen in the hepatocellular parenchyma?\\
\toprule[1pt]
\multirow{8}{*}{\textbf{TUM-Uteria}}&hemorrhagic change&Tissue-level&What hemorrhagic change is described in the specimen?\\
&stromal collagen&Tissue-level&What collagenous or fibrotic change is described in the stroma?\\
&necrosis&Tissue-level&What necrotic change is described in the lesion?\\
&smooth muscle architecture&Architectural-level&What architectural pattern is formed by the smooth-muscle spindle cells?\\
&invasion&Architectural-level&Which tissue or compartment is invaded by the lesion?\\
&epithelial maturation&Architectural-level&How is maturation of the squamous epithelium described?\\
&smooth muscle cytology&Cellular-level&How are the smooth-muscle spindle cells morphologically described?\\
&mitotic activity&Cellular-level&How is mitotic activity described in the lesional cells?\\

\bottomrule[1.5pt]

\end{tabular}

\end{lrbox}
\scalebox{0.79}{\usebox{\tablebox}}
\end{table*}

\subsection{Quantitative QA Performance of PathPilot}
\label{sec:qa_exp}
We evaluate PathPilot through end-to-end open-ended QA under both internal and external validation settings. In each setting, performance is reported for tissue-level QA performance, architectural-level QA performance, and cellular-level  QA performance according to the focus of each question, together with the overall QA performance averaged across all questions.

\subsubsection{Internal Validation.}

We first evaluate PathPilot on the TUM-Uteria and TUM Liver internal cohorts. As shown in Fig.~\ref{fig:performance} and Tables~\ref{results:tum-uteria}--\ref{results:tum-liver}, the results are examined in terms of overall QA performance, followed by tissue-level, architectural-level, and cellular-level QA performance to assess PathPilot across different clinically relevant aspects of pathological assessment.

\textbf{Average QA performance.}
As summarized in Fig.~\ref{fig:performance} and Tables~\ref{results:tum-uteria} and \ref{results:tum-liver}, PathPilot achieves the \textbf{strongest average QA performance} on both internal validation cohorts, with the \textbf{highest GACC and PBR and the lowest IAR}. In terms of GACC, PathPilot exceeds the strongest baseline by 5.75 \% on TUM-Uteria and 11.17 \% on TUM Liver.

Across the five baseline categories, the strongest general-purpose VLMs achieve GACC values of 38.32\% on TUM-Uteria and 40.00\% on TUM Liver, while the strongest general medical VLMs reach 44.07\% and 46.98\%, respectively. Pathology-specialized thumbnail-based VLMs are substantially more competitive, with PathGen-LLaVA reaching 54.61\% and 48.60\%, while the strongest patch-based WSI VLMs achieve 52.99\% and 44.88\%. Among existing pathology agents, PathNavigate achieves 46.35\% and 49.30\% on the two cohorts. In comparison, PathPilot achieves 60.36\% and 60.47\%. The substantially weaker performance of general-purpose and general medical VLMs indicates that broad visual-language or medical knowledge alone is insufficient for effective WSI question answering. We therefore focus the subsequent category-specific comparisons on pathology-specialized thumbnail VLMs, patch-based WSI VLMs, and pathology agents, which provide more relevant baselines for comparing different forms of WSI evidence access and navigation.

\begin{table*}[t]
\centering
\setlength{\fboxsep}{0pt}

\caption{QA Performance ($\%$) at Internal TUM-Uteria dataset  \cite{li2026paired}. \colorbox{first}{Best} and \colorbox{second}{second-best} results are highlighted. Baselines are grouped from top to bottom as natural-image-based general-purpose VLMs, general medical VLMs, pathology-specialized thumbnail-based VLMs, patch-based WSI VLMs, and pathology agents for open-ended WSI question answering. All methods use deterministic decoding with a temperature of 0.  Five repeated runs of PathPilot produced identical results, yielding zero run-to-run variance.} \label{results:tum-uteria}
\begin{lrbox}{\tablebox}
\begin{tabular}{lcccccccccccc}
\toprule[1.5pt]
\toprule[1.5pt]
\multicolumn{1}{c}{\multirow{2}{*}{\textbf{Methods}}} & \multicolumn{3}{c}{ \shortstack{\textbf{Average}\\[-0.45ex]\rule{3.25cm}{0.9pt}}} & \multicolumn{3}{c}{ \shortstack{\textbf{Tissue-level}\\[-0.45ex]\rule{3.25cm}{0.9pt}}} & \multicolumn{3}{c}{ \shortstack{\textbf{Architectural-level}\\[-0.45ex]\rule{3.25cm}{0.9pt}}} & \multicolumn{3}{c}{ \shortstack{\textbf{Cellular-level}\\[-0.45ex]\rule{3.25cm}{0.9pt}}} \\[-0.35ex] \multicolumn{1}{c}{} & \multicolumn{1}{c}{\textbf{GACC. $\uparrow$
}} & \multicolumn{1}{c}{\textbf{PBR. $\uparrow$}} & \multicolumn{1}{c}{\textbf{IAR. $\downarrow$}} & \multicolumn{1}{c}{\textbf{GACC. $\uparrow$}} & \multicolumn{1}{c}{\textbf{PBR. $\uparrow$}} & \multicolumn{1}{c}{\textbf{IAR. $\downarrow$}} & \multicolumn{1}{c}{\textbf{GACC. $\uparrow$}} & \multicolumn{1}{c}{\textbf{PBR. $\uparrow$}} & \multicolumn{1}{c}{\textbf{IAR. $\downarrow$}} & \multicolumn{1}{c}{\textbf{GACC. $\uparrow$}} & \multicolumn{1}{c}{\textbf{PBR. $\uparrow$}} & \multicolumn{1}{c}{\textbf{IAR. $\downarrow$}} \\
\toprule[1pt]
Gemma-3-12b-it \cite{Kamath2025Gemma3T}&17.49\%&12.57\%&71.86\%&6.04\%&7.69\%&92.31\%&29.64\%&25.00\%&57.14\%&32.08\%&14.58\%&41.67\%\\
Qwen-2.5-VL-7B-Instruct \cite{bai2025qwen25vl}&34.37\%&13.77\%&36.53\%&40.44\%&16.48\%&25.27\%&26.79\%&17.86\%&53.57\%&27.29\%&6.25\%&47.92\%\\
Qwen-2.5-VL-32B-Instruct \cite{bai2025qwen25vl}&38.32\%&16.77\%&30.54\%&46.70\%&23.08\%&16.48\%&22.14\%&10.71\%&60.71\%&31.88\%&8.33\%&39.58\%\\
Qwen-3-VL-8B-Instruct \cite{bai2025qwen3}&16.17\%&16.77\%&75.45\%&4.51\%&4.40\%&93.41\%&22.86\%&28.57\%&67.86\%&34.38\%&33.33\%&45.83\%\\
Qwen-3-VL-32B-Instruct \cite{bai2025qwen3}&17.19\%&19.76\%&74.25\%&4.51\%&4.40\%&93.41\%&25.71\%&25.00\%&60.71\%&36.25\%&45.83\%&45.83\%\\
InternVL3-8B-Instruct \cite{zhu2025internvl3}&15.87\%&8.98\%&71.86\%&16.81\%&9.89\%&70.33\%&18.93\%&14.29\%&67.86\%&12.29\%&4.17\%&77.08\%\\
InternVL3-14B-Instruct \cite{zhu2025internvl3}&17.49\%&8.38\%&69.46\%&28.02\%&14.29\%&51.65\%&2.50\%&3.57\%&96.43\%&6.25\%&0.00\%&87.50\%\\
InternVL3.5-8B \cite{wang2025internvl3}&16.89\%&11.38\%&71.86\%&26.37\%&19.78\%&57.14\%&11.43\%&3.57\%&78.57\%&2.08\%&0.00\%&95.83\%\\
InternVL3.5-14B \cite{wang2025internvl3}&18.68\%&10.18\%&67.07\%&29.67\%&17.58\%&48.35\%&3.57\%&0.00\%&92.86\%&6.67\%&2.08\%&87.50\%\\
\midrule

LLaVA-Med-v1.5 \cite{li2023llava}&21.62\%&17.96\%&66.47\%&17.47\%&13.19\%&70.33\%&15.71\%&7.14\%&71.43\%&32.92\%&33.33\%&56.25\%\\
HuatuoGPT-Vision-7B \cite{chen2024towardsHuatuoGPT}&35.51\%&17.37\%&35.93\%&48.24\%&24.18\%&13.19\%&10.36\%&7.14\%&82.14\%&26.04\%&10.42\%&52.08\%\\
Med-Flamingo-9B \cite{moor2023med}&24.55\%&2.99\%&52.10\%&12.97\%&4.40\%&75.82\%&25.71\%&3.57\%&50.00\%&45.83\%&0.00\%&\colorbox{second}{8.33\%}\\
Lingshu-I-8B \cite{xu2025lingshu}&33.47\%&20.36\%&41.92\%&42.97\%&31.87\%&27.47\%&15.71\%&10.71\%&{75.00\%}&25.83\%&4.17\%&50.00\%\\
MedGemma-4b-it \cite{sellergren2025medgemma}&29.64\%&20.96\%&49.10\%&39.78\%&34.07\%&34.07\%&10.36\%&7.14\%&82.14\%&21.67\%&4.17\%&58.33\%\\
MedGemma-27b-it \cite{sellergren2025medgemma}&26.83\%&23.35\%&57.49\%&11.43\%&7.69\%&80.22\%&59.64\%&\colorbox{first}{75.00\%}&21.43\%&36.88\%&22.92\%&35.42\%\\

HealthGPT-Pro-4B  \cite{lin2025healthgptmedicallargevisionlanguage}&41.02\%&23.35\%&28.14\%&49.67\%&26.37\%&12.09\%&20.36\%&3.57\%&60.71\%&36.67\%&29.17\%&39.58\%\\
HealthGPT-Pro-8B  \cite{lin2025healthgptmedicallargevisionlanguage}&44.07\% &25.75\% &20.96\% & 49.23\%&29.67\%&12.09\%&34.64\%&21.43\%&39.29\%&39.79\%&20.83\%&27.08\% \\
\midrule

Hulu-Med-7B \cite{jiang2025hulu}&40.12\% &26.95\% &33.53\%&39.56\%&26.37\%&34.07\%&53.21\%&42.86\%&17.86\%&33.54\%&18.75\%&41.67\%\\
Hulu-Med-14B \cite{jiang2025hulu}&37.96\%&25.75\%&36.53\%&38.13\%&26.37\%&36.26\%&50.36\%&39.29\%&21.43\%&30.42\%&16.67\%&45.83\%\\
PathGen-LLaVA \cite{sun2025pathgen}&\colorbox{second}{54.61\%}&\colorbox{second}{47.31\%}&14.37\%&\colorbox{second}{52.20\%}&35.16\%&\colorbox{second}{10.99\%}&58.57\%&\colorbox{second}{64.29\%}&21.43\%&\colorbox{second}{56.88\%}&\colorbox{second}{60.42\%}&16.67\%\\
Patho-R1-7B \cite{zhang2026patho}&35.63\%&34.13\%&43.11\%&27.14\%&28.57\%&57.14\%&\colorbox{second}{60.00\%}&\colorbox{first}{75.00\%}&\colorbox{second}{14.29\%}&37.50\%&20.83\%&33.33\%\\
Quilt-LLaVA-v1.5-7b \cite{seyfioglu2024quilt}&40.96\%&17.96\%&23.95\%&48.35\%&18.68\%&\colorbox{first}{7.69\%}&20.00\%&10.71\%&64.29\%&39.17\%&20.83\%&31.25\%\\
\midrule

WSI-VQA \cite{chen2024wsi}&7.78\%&2.99\%&85.63\%&10.99\%&5.49\%&80.22\%&0.00\%&0.00\%&100.00\%&6.25\%&0.00\%&87.50\%\\
PA-LLaVA \cite{dai2024pa}&8.38\%&2.99\%&84.43\%&12.09\%&5.49\%&78.02\%&0.00\%&0.00\%&100.00\%&6.25\%&0.00\%&87.50\%\\
WSI-LLaVA \cite{liang2025wsi}&52.99\%&43.11\%&\colorbox{second}{13.77\%}&50.11\%&\colorbox{second}{36.26\%}&14.29\%&56.07\%&57.14\%&21.43\%&56.67\%&47.92\%&\colorbox{second}{8.33\%}\\
MedGemma-1.5-4b-it \cite{sellergren2026medgemma}&31.92\%&25.15\%&47.31\%&31.54\%&31.87\%&51.65\%&29.64\%&14.29\%&46.43\%&33.96\%&18.75\%&39.58\%\\
SlideChat \cite{chen2025slidechat}&30.96\%&23.35\%&48.50\%&30.77\%&30.77\%&52.75\%&27.86\%&14.29\%&50.00\%&33.13\%&14.58\%&39.58\%\\
ANTONI-Alpha \cite{moonemans2025democratizing}&33.59\%&27.54\%&44.91\%&33.08\%&34.07\%&49.45\%&33.21\%&14.29\%&39.29\%&34.79\%&22.92\%&39.58\%\\
\midrule
PathAgent \cite{chen2025pathagent}&45.21\%&34.13\%&23.95\%&44.29\%&31.87\%&24.18\%&48.93\%&46.43\%&25.00\%&44.79\%&31.25\%&22.92\%\\
PathNavigate \cite{yang2026pathnavigate}&46.35\%&32.34\%&22.75\%&45.27\%&34.07\%&23.08\%&45.71\%&35.71\%&35.71\%&48.75\%&27.08\%&14.58\%\\

Ours&\colorbox{first}{60.36\%}&\colorbox{first}{61.08\%} &\colorbox{first}{6.59\%}&\colorbox{first}{57.69\%}&\colorbox{first}{52.75\%}&\colorbox{first}{7.69\%}&\colorbox{first}{65.00\%}&\colorbox{first}{75.00\%}&\colorbox{first}{10.71\%}&\colorbox{first}{62.71\%}&\colorbox{first}{68.75\%}&\colorbox{first}{2.08\%}\\

\bottomrule[1.5pt]

\end{tabular}

\end{lrbox}
\scalebox{0.7}{\usebox{\tablebox}}
\end{table*}

\textbf{Tissue-level QA performance.}
Tissue-level questions require broader spatial evidence about the presence, extent, and distribution of pathological alterations across the WSI. Because these findings may be heterogeneous or distributed across multiple tissue compartments, reliable answering requires both broader tissue context and the identification and integration of relevant evidence \textbf{across spatial regions}.

As shown in Fig.~\ref{fig:performance} and Tables~\ref{results:tum-uteria} and \ref{results:tum-liver}, PathPilot achieves the \textbf{strongest tissue-level QA performance} on both internal cohorts. Compared with the strongest GACC baseline, it improves performance by 5.49 \% on TUM-Uteria and 10.00 \% on TUM Liver. It also achieves the highest PBR and lowest IAR on both cohorts, indicating \textbf{more highly correct responses and fewer complete errors.}

The comparison across WSI analysis paradigms further illustrates the effect of different evidence-acquisition strategies. Among thumbnail-based pathology VLMs, PathGen-LLaVA achieves the strongest tissue-level GACC of 52.20\% on TUM-Uteria, while Patho-R1-7B achieves 47.78\% on TUM Liver; although these models preserve broad slide context, a single downsampled view limits access to spatially heterogeneous findings. Among patch-based WSI VLMs, WSI-LLaVA achieves 50.11\% and 53.33\%, respectively, retaining richer local information but relying on predefined, question-independent sampling. Among existing pathology agents, PathNavigate achieves 45.27\% and 48.89\%, respectively, but uses more constrained region or magnification strategies. In contrast, PathPilot combines CrossMagLoc-based candidate localization with EviRegionSearch-based evidence-adaptive exploration, enabling spatially distributed findings to be progressively identified and integrated according to the evidence required by the current question.

\begin{table*}[t]
\centering
\setlength{\fboxsep}{0pt}

\caption{QA Performance ($\%$) at Internal TUM Liver dataset. \colorbox{first}{Best} and \colorbox{second}{second-best} results are highlighted. Baselines are grouped from top to bottom as natural-image-based general-purpose VLMs, general medical VLMs, pathology-specialized thumbnail-based VLMs, patch-based WSI VLMs, and pathology agents for open-ended WSI question answering. All methods use deterministic decoding with a temperature of 0. Five repeated runs of ours produced identical results, yielding zero run-to-run variance.} \label{results:tum-liver}
\begin{lrbox}{\tablebox}
\begin{tabular}{lcccccccccccc}
\toprule[1.5pt]
\toprule[1.5pt]
\multicolumn{1}{c}{\multirow{2}{*}{\textbf{Methods}}} & \multicolumn{3}{c}{ \shortstack{\textbf{Average}\\[-0.45ex]\rule{3.25cm}{0.9pt}}} & \multicolumn{3}{c}{ \shortstack{\textbf{Tissue-level}\\[-0.45ex]\rule{3.25cm}{0.9pt}}} & \multicolumn{3}{c}{ \shortstack{\textbf{Architectural-level}\\[-0.45ex]\rule{3.25cm}{0.9pt}}} & \multicolumn{3}{c}{ \shortstack{\textbf{Cellular-level}\\[-0.45ex]\rule{3.25cm}{0.9pt}}} \\[-0.35ex] \multicolumn{1}{c}{} & \multicolumn{1}{c}{\textbf{GACC. $\uparrow$
}} & \multicolumn{1}{c}{\textbf{PBR. $\uparrow$}} & \multicolumn{1}{c}{\textbf{IAR. $\downarrow$}} & \multicolumn{1}{c}{\textbf{GACC. $\uparrow$}} & \multicolumn{1}{c}{\textbf{PBR. $\uparrow$}} & \multicolumn{1}{c}{\textbf{IAR. $\downarrow$}} & \multicolumn{1}{c}{\textbf{GACC. $\uparrow$}} & \multicolumn{1}{c}{\textbf{PBR. $\uparrow$}} & \multicolumn{1}{c}{\textbf{IAR. $\downarrow$}} & \multicolumn{1}{c}{\textbf{GACC. $\uparrow$}} & \multicolumn{1}{c}{\textbf{PBR. $\uparrow$}} & \multicolumn{1}{c}{\textbf{IAR. $\downarrow$}} \\
\toprule[1pt]
Gemma-3-12b-it \cite{Kamath2025Gemma3T}&35.12\%&25.58\%&41.86\%&32.22\%&22.22\%&44.44\%&33.00\%&0.00\%&30.00\%&37.08\%&37.50\%&45.83\%\\
Qwen-2.5-VL-7B-Instruct \cite{bai2025qwen25vl}&36.98\%&13.95\%&30.23\%&43.33\%&22.22\%&22.22\%&40.00\%&20.00\%&30.00\%&33.33\%&8.33\%&33.33\%\\
Qwen-2.5-VL-32B-Instruct \cite{bai2025qwen25vl}&40.00\%&16.28\%&27.91\%&43.33\%&22.22\%&22.22\%&35.00\%&0.00\%&30.00\%&40.83\%&20.83\%&29.17\%\\
Qwen-3-VL-8B-Instruct \cite{bai2025qwen3}&3.95\%&2.33\%&93.02\%&11.11\%&0.00\%&77.78\%&0.00\%&0.00\%&100.00\%&2.92\%&4.17\%&95.83\%\\
Qwen-3-VL-32B-Instruct \cite{bai2025qwen3}&24.88\%&20.93\%&69.77\%&16.67\%&11.11\%&77.78\%&32.00\%&30.00\%&60.00\%&25.00\%&20.83\%&70.83\%\\
InternVL3-8B-Instruct \cite{zhu2025internvl3}&29.07\%&23.26\%&51.16\%&32.22\%&22.22\%&44.44\%&59.00\%&\colorbox{second}{70.00\%}&\colorbox{second}{10.00\%}&15.42\%&4.17\%&70.83\%\\
InternVL3-14B-Instruct \cite{zhu2025internvl3}&18.14\%&9.30\%&67.44\%&35.56\%&11.11\%&33.33\%&26.00\%&30.00\%&60.00\%&8.33\%&0.00\%&83.33\%\\
InternVL3.5-8B \cite{wang2025internvl3}&34.19\%&20.93\%&44.19\%&35.56\%&11.11\%&33.33\%&58.00\%&\colorbox{second}{70.00\%}&30.00\%&23.75\%&4.17\%&54.17\%\\
InternVL3.5-14B \cite{wang2025internvl3}&31.16\%&18.60\%&46.51\%&45.56\%&33.33\%&22.22\%&48.00\%&50.00\%&30.00\%&18.75\%&0.00\%&62.50\%\\
\midrule

LLaVA-Med-v1.5 \cite{li2023llava}&33.95\%&20.93\%&41.86\%&37.78\%&22.22\%&33.33\%&48.00\%&40.00\%&20.00\%&26.67\%&12.50\%&54.17\%\\
HuatuoGPT-Vision-7B \cite{chen2024towardsHuatuoGPT}&42.79\%&27.91\%&25.58\%&41.11\%&11.11\%&22.22\%&55.00\%&50.00\%&\colorbox{second}{10.00\%}&38.33\%&25.00\%&33.33\%\\
Med-Flamingo-9B \cite{moor2023med}&28.84\%&6.98\%&44.19\%&22.22\%&11.11\%&55.56\%&40.00\%&0.00\%&20.00\%&26.67\%&8.33\%&50.00\%\\
Lingshu-I-8B \cite{xu2025lingshu}&42.33\%&25.58\%&30.23\%&40.00\%&44.44\%&33.33\%&47.00\%&40.00\%&40.00\%&41.25\%&12.50\%&25.00\%\\
MedGemma-4b-it \cite{sellergren2025medgemma}&44.42\%&20.93\%&\colorbox{second}{20.93\%}&45.56\%&33.33\%&22.22\%&47.00\%&10.00\%&\colorbox{second}{10.00\%}&42.92\%&20.83\%&25.00\%\\
MedGemma-27b-it \cite{sellergren2025medgemma}&35.58\%&13.95\%&34.88\%&33.33\%&11.11\%&44.44\%&49.00\%&20.00\%&\colorbox{second}{10.00\%}&30.83\%&12.50\%&41.67\%\\

HealthGPT-Pro-4B  \cite{lin2025healthgptmedicallargevisionlanguage}&43.02\%&30.23\%&30.23\%&42.22\%&44.44\%&33.33\%&39.00\%&30.00\%&40.00\%&45.00\%&25.00\%&25.00\%\\
HealthGPT-Pro-8B  \cite{lin2025healthgptmedicallargevisionlanguage}&32.09\%&23.26\%&46.51\%&42.22\%&44.44\%&33.33\%&31.00\%&30.00\%&50.00\%&28.75\%&12.50\%&50.00\% \\
\midrule

Hulu-Med-7B \cite{jiang2025hulu}&37.67\%&27.91\%&37.21\%&40.00\%&33.33\%&33.33\%&27.00\%&10.00\%&50.00\%&41.25\%&33.33\%&33.33\%\\
Hulu-Med-14B \cite{jiang2025hulu}&41.16\%&32.56\%&34.88\%&37.78\%&22.22\%&33.33\%&39.00\%&30.00\%&40.00\%&43.33\%&37.50\%&33.33\%\\
PathGen-LLaVA \cite{sun2025pathgen}&48.60\%&37.21\%&23.26\%&45.56\%&33.33\%&22.22\%&56.00\%&60.00\%&30.00\%&46.67\%&29.17\%&{20.83\%}\\
Patho-R1-7B \cite{zhang2026patho}&44.42\%&34.88\%&34.88\%&47.78\%&\colorbox{second}{55.56\%}&33.33\%&20.00\%&10.00\%&70.00\%&\colorbox{second}{53.33\%}&37.50\%&{20.83\%}\\
Quilt-LLaVA-v1.5-7b \cite{seyfioglu2024quilt}&40.00\%&16.28\%&27.91\%&30.00\%&11.11\%&44.44\%&32.00\%&10.00\%&40.00\%&47.08\%&20.83\%&\colorbox{second}{16.67\%}\\

\midrule

WSI-VQA \cite{chen2024wsi}&14.88\%&4.65\%&72.09\%&5.56\%&0.00\%&88.89\%&29.00\%&20.00\%&50.00\%&12.50\%&0.00\%&75.00\%\\
PA-LLaVA \cite{dai2024pa}&16.51\%&6.98\%&69.77\%&5.56\%&0.00\%&88.89\%&26.00\%&30.00\%&60.00\%&16.67\%&0.00\%&66.67\%\\
WSI-LLaVA \cite{liang2025wsi}&43.02\%&\colorbox{second}{39.53\%}&32.56\%&\colorbox{second}{53.33\%}&44.44\%&\colorbox{second}{11.11\%}&24.00\%&30.00\%&70.00\%&47.08\%&\colorbox{second}{41.67\%}&25.00\%\\
MedGemma-1.5-4b-it \cite{sellergren2026medgemma}&44.88\%&23.26\%&\colorbox{second}{20.93\%}&45.56\%&33.33\%&22.22\%&49.00\%&20.00\%&\colorbox{second}{10.00\%}&42.92\%&20.83\%&25.00\%\\
SlideChat \cite{chen2025slidechat}&41.63\%&27.91\%&27.91\%&37.78\%&22.22\%&33.33\%&\colorbox{second}{64.00\%}&\colorbox{second}{70.00\%}&\colorbox{first}{0.00\%}&33.75\%&12.50\%&37.50\%\\
ANTONI-Alpha \cite{moonemans2025democratizing}&43.26\%&27.91\%&30.23\%&51.11\%&44.44\%&22.22\%&44.00\%&40.00\%&40.00\%&40.00\%&16.67\%&29.17\%\\
\midrule
PathAgent \cite{chen2025pathagent}&44.19\%&34.88\%&32.56\%&43.33\%&22.22\%&22.22\%&56.00\%&60.00\%&30.00\%&39.58\%&29.17\%&37.50\%\\
PathNavigate \cite{yang2026pathnavigate}&\colorbox{second}{49.30\%}&37.21\%&23.26\%&48.89\%&22.22\%&\colorbox{second}{11.11\%}&63.00\%&\colorbox{second}{70.00\%}&20.00\%&43.75\%&29.17\%&29.17\%\\

Ours&\colorbox{first}{60.47\%}&\colorbox{first}{62.79\%} &\colorbox{first}{6.98\%}&\colorbox{first}{63.33\%}&\colorbox{first}{66.67\%}&\colorbox{first}{0.00\%}&\colorbox{first}{69.00\%}&\colorbox{first}{90.00\%}&\colorbox{second}{10.00\%}&\colorbox{first}{55.83\%}&\colorbox{first}{50.00\%}&\colorbox{first}{8.33\%}\\

\bottomrule[1.5pt]

\end{tabular}

\end{lrbox}
\scalebox{0.7}{\usebox{\tablebox}}
\end{table*}

\begin{table*}[t]
\centering
\setlength{\fboxsep}{0pt}

\caption{QA Performance ($\%$) at External TUM Skin dataset. \colorbox{first}{Best} and \colorbox{second}{second-best} results are highlighted. Baselines are grouped from top to bottom as natural-image-based general-purpose VLMs, general medical VLMs, pathology-specialized thumbnail-based VLMs, patch-based WSI VLMs, and pathology agents for open-ended WSI question answering. All methods use deterministic decoding with a temperature of 0.  Five repeated runs of ours produced identical results, yielding zero run-to-run variance.} 
\label{results:tum-skin}
\begin{lrbox}{\tablebox}
\begin{tabular}{lcccccccccccc}
\toprule[1.5pt]
\toprule[1.5pt]
\multicolumn{1}{c}{\multirow{2}{*}{\textbf{Methods}}} & \multicolumn{3}{c}{ \shortstack{\textbf{Average}\\[-0.45ex]\rule{3.25cm}{0.9pt}}} & \multicolumn{3}{c}{ \shortstack{\textbf{Tissue-level}\\[-0.45ex]\rule{3.25cm}{0.9pt}}} & \multicolumn{3}{c}{ \shortstack{\textbf{Architectural-level}\\[-0.45ex]\rule{3.25cm}{0.9pt}}} & \multicolumn{3}{c}{ \shortstack{\textbf{Cellular-level}\\[-0.45ex]\rule{3.25cm}{0.9pt}}} \\[-0.35ex] \multicolumn{1}{c}{} & \multicolumn{1}{c}{\textbf{GACC. $\uparrow$
}} & \multicolumn{1}{c}{\textbf{PBR. $\uparrow$}} & \multicolumn{1}{c}{\textbf{IAR. $\downarrow$}} & \multicolumn{1}{c}{\textbf{GACC. $\uparrow$}} & \multicolumn{1}{c}{\textbf{PBR. $\uparrow$}} & \multicolumn{1}{c}{\textbf{IAR. $\downarrow$}} & \multicolumn{1}{c}{\textbf{GACC. $\uparrow$}} & \multicolumn{1}{c}{\textbf{PBR. $\uparrow$}} & \multicolumn{1}{c}{\textbf{IAR. $\downarrow$}} & \multicolumn{1}{c}{\textbf{GACC. $\uparrow$}} & \multicolumn{1}{c}{\textbf{PBR. $\uparrow$}} & \multicolumn{1}{c}{\textbf{IAR. $\downarrow$}} \\
\toprule[1pt]
Gemma-3-12b-it  \cite{Kamath2025Gemma3T}&18.33\%&11.36\%&69.70\%&9.04\%&9.59\%&89.04\%&33.44\%&18.75\%&40.63\%&25.56\%&7.41\%&51.85\%\\
Qwen-2.5-VL-7B-Instruct  \cite{bai2025qwen25vl}&35.00\%&15.15\%&36.36\%&37.53\%&21.92\%&34.25\%&20.94\%&3.13\%&59.38\%&44.81\%&11.11\%&14.81\%\\
Qwen-2.5-VL-32B-Instruct  \cite{bai2025qwen25vl}&43.86\%&17.42\%&\colorbox{second}{21.21\%}&45.75\%&23.29\%&21.92\%&35.00\%&3.13\%&31.25\%&49.26\%&18.52\%&\colorbox{second}{7.41\%}\\
Qwen-3-VL-8B-Instruct  \cite{bai2025qwen3}&31.82\%&19.70\%&49.24\%&30.27\%&26.03\%&58.90\%&22.50\%&3.13\%&56.25\%&47.04\%&22.22\%&14.81\%\\
Qwen-3-VL-32B-Instruct  \cite{bai2025qwen3}&33.11\%&19.70\%&46.21\%&31.64\%&27.40\%&57.53\%&25.94\%&3.13\%&46.88\%&45.56\%&18.52\%&14.81\%\\
InternVL3-8B-Instruct  \cite{zhu2025internvl3}&15.53\%&4.55\%&71.21\%&19.86\%&8.22\%&64.38\%&4.69\%&0.00\%&90.63\%&16.67\%&0.00\%&66.67\%\\
InternVL3-14B-Instruct  \cite{zhu2025internvl3}&10.91\%&1.52\%&78.79\%&13.56\%&2.74\%&73.97\%&3.13\%&0.00\%&93.75\%&12.96\%&0.00\%&74.07\%\\
InternVL3.5-8B  \cite{wang2025internvl3}&17.05\%&5.30\%&68.94\%&19.32\%&5.48\%&64.38\%&6.25\%&0.00\%&87.50\%&23.70\%&11.11\%&59.26\%\\
InternVL3.5-14B  \cite{wang2025internvl3}&18.18\%&4.55\%&65.91\%&15.34\%&1.37\%&69.86\%&14.06\%&3.13\%&75.00\%&30.74\%&14.81\%&44.44\%\\
\midrule

LLaVA-Med-v1.5  \cite{li2023llava}&37.12\%&31.82\%&43.94\%&47.53\%&\colorbox{second}{50.68\%}&34.25\%&17.50\%&9.38\%&68.75\%&32.22\%&7.41\%&40.74\%\\
HuatuoGPT-Vision-7B  \cite{chen2024towardsHuatuoGPT}&42.88\%&23.48\%&25.00\%&45.89\%&21.92\%&\colorbox{second}{17.81\%}&31.25\%&21.88\%&50.00\%&48.52\%&29.63\%&14.81\%\\
Med-Flamingo-9B  \cite{moor2023med}&17.58\%&4.55\%&66.67\%&6.71\%&2.74\%&87.67\%&27.50\%&12.50\%&50.00\%&35.19\%&0.00\%&29.63\%\\
Lingshu-I-8B  \cite{xu2025lingshu}&40.91\%&27.27\%&29.55\%&44.38\%&31.51\%&24.66\%&30.00\%&25.00\%&50.00\%&44.44\%&18.52\%&18.52\%\\
MedGemma-4b-it  \cite{sellergren2025medgemma}&36.36\%&19.70\%&40.15\%&38.77\%&34.25\%&45.21\%&26.56\%&0.00\%&46.88\%&41.48\%&3.70\%&18.52\%\\
MedGemma-27b-it  \cite{sellergren2025medgemma}&42.95\%&29.55\%&31.82\%&46.03\%&39.73\%&32.88\%&29.06\%&15.63\%&50.00\%&51.11\%&18.52\%&\colorbox{second}{7.41\%}\\

HealthGPT-Pro-4B  \cite{lin2025healthgptmedicallargevisionlanguage}&47.65\% &31.82\%&28.79\%&53.97\%&39.73\%&26.03\%&38.13\%&28.13\%&40.63\%&41.85\%&14.81\%&22.22\% \\
HealthGPT-Pro-8B  \cite{lin2025healthgptmedicallargevisionlanguage}&41.36\%&28.03\% &32.58\%&49.18\%&41.10\%&24.66\%&20.00\%&9.38\%&65.63\%&45.56\%&14.81\%&14.81\%\\
\midrule

Hulu-Med-7B  \cite{jiang2025hulu}&44.47\%&29.55\%&28.79\%&50.14\%&41.10\%&26.03\%&26.88\%&12.50\%&53.13\%&50.00\%&18.52\%&\colorbox{second}{7.41\%}\\
Hulu-Med-14B  \cite{jiang2025hulu}&38.64\%&31.06\%&40.15\%&41.78\%&39.73\%&39.73\%&20.63\%&15.63\%&68.75\%&\colorbox{second}{51.48\%}&25.93\%&\colorbox{second}{7.41\%}\\
PathGen-LLaVA  \cite{sun2025pathgen}&42.58\%&33.33\%&31.82\%&\colorbox{second}{54.52\%}&\colorbox{second}{50.68\%}&\colorbox{second}{17.81\%}&22.81\%&12.50\%&59.38\%&33.70\%&11.11\%&37.04\%\\
Patho-R1-7B  \cite{zhang2026patho}&\colorbox{second}{47.95\%}&37.12\%&23.48\%&53.29\%&49.32\%&20.55\%&34.38\%&15.63\%&37.50\%&49.63\%&29.63\%&14.81\%\\
Quilt-LLaVA-v1.5-7b  \cite{seyfioglu2024quilt}&16.97\%&5.30\%&68.18\%&9.86\%&1.37\%&80.82\%&25.31\%&9.38\%&53.13\%&26.30\%&11.11\%&51.85\%\\
\midrule
WSI-VQA  \cite{chen2024wsi}&14.70\%&12.88\%&78.03\%&24.93\%&21.92\%&63.01\%&2.19\%&3.13\%&96.88\%&1.85\%&0.00\%&96.30\%\\
PA-LLaVA  \cite{dai2024pa}&16.14\%&14.39\%&75.76\%&26.85\%&24.66\%&60.27\%&3.75\%&3.13\%&93.75\%&1.85\%&0.00\%&96.30\%\\
WSI-LLaVA  \cite{liang2025wsi}&38.48\%&25.00\%&37.12\%&46.85\%&41.10\%&30.14\%&20.00\%&6.25\%&62.50\%&37.78\%&3.70\%&25.93\%\\
MedGemma-1.5-4b-it  \cite{sellergren2026medgemma}&37.12\%&28.79\%&43.18\%&38.08\%&36.99\%&49.32\%&28.13\%&15.63\%&50.00\%&45.19\%&22.22\%&18.52\%\\
SlideChat  \cite{chen2025slidechat}&37.95\%&31.06\%&42.42\%&38.36\%&38.36\%&49.32\%&32.50\%&21.88\%&43.75\%&43.33\%&22.22\%&22.22\%\\
ANTONI-Alpha  \cite{moonemans2025democratizing}&43.41\%&31.82\%&27.27\%&42.47\%&31.51\%&30.14\%&\colorbox{second}{45.00\%}&37.50\%&\colorbox{second}{25.00\%}&44.07\%&25.93\%&22.22\%\\
\midrule
PathAgent  \cite{chen2025pathagent}&42.65\%&\colorbox{second}{42.42\%}&32.58\%&44.66\%&46.58\%&30.14\%&33.75\%&31.25\%&46.88\%&47.78\%&\colorbox{second}{44.44\%}&22.22\%\\
PathNavigate  \cite{yang2026pathnavigate}&45.00\%&37.88\%&28.79\%&45.07\%&36.99\%&28.77\%&42.19\%&\colorbox{second}{40.63\%}&37.50\%&48.15\%&37.04\%&18.52\%\\

Ours&\colorbox{first}{61.59\%}&\colorbox{first}{62.12\%}&\colorbox{first}{6.06\%} &\colorbox{first}{65.21\%}&\colorbox{first}{65.75\%}&\colorbox{first}{4.11\%}&\colorbox{first}{53.75\%}&\colorbox{first}{56.25\%}&\colorbox{first}{12.50\%}&\colorbox{first}{61.11\%}&\colorbox{first}{59.26\%}&\colorbox{first}{3.70\%}\\

\bottomrule[1.5pt]

\end{tabular}

\end{lrbox}
\scalebox{0.7}{\usebox{\tablebox}}
\end{table*}

\begin{table*}[t]
\centering
\setlength{\fboxsep}{0pt}
\setlength{\tabcolsep}{2.8pt}

\caption{
QA performance ($\%$) on the external WSI-Bench TCGA-SKCM and TCGA-LIHC datasets~ \cite{liang2025wsi}.
\colorbox{first}{Best} and \colorbox{second}{second-best} results are highlighted.
Baselines are grouped from top to bottom as natural-image-based general-purpose VLMs, general medical VLMs, pathology-specialized thumbnail-based VLMs, patch-based WSI VLMs, and pathology agents for open-ended WSI question answering.
All methods use deterministic decoding with a temperature of 0.
Five repeated runs of ours produced identical results, yielding zero run-to-run variance.
}
\label{results:tcga_external}

\begin{lrbox}{\tablebox}
\begin{tabular}{lcccccccccccc}

\toprule[1.5pt]

\multirow{3}{*}{\textbf{Methods}}
&
\multicolumn{6}{c}{\textbf{TCGA-SKCM}}
&
\multicolumn{6}{c}{\textbf{TCGA-LIHC}}
\\

\cmidrule(lr){2-7}
\cmidrule(lr){8-13}

&
\multicolumn{3}{c}{\textbf{Morphology}}
&
\multicolumn{3}{c}{\textbf{All QA (supporting)}}&
\multicolumn{3}{c}{\textbf{Morphology}}
&
\multicolumn{3}{c}{\textbf{All QA (supporting)}}\\

\cmidrule(lr){2-4}
\cmidrule(lr){5-7}
\cmidrule(lr){8-10}
\cmidrule(lr){11-13}

&
\textbf{GACC. $\uparrow$}
&
\textbf{PBR. $\uparrow$}
&
\textbf{IAR. $\downarrow$}
&
\textbf{GACC. $\uparrow$}
&
\textbf{PBR. $\uparrow$}
&
\textbf{IAR. $\downarrow$}
&
\textbf{GACC. $\uparrow$}
&
\textbf{PBR. $\uparrow$}
&
\textbf{IAR. $\downarrow$}
&
\textbf{GACC. $\uparrow$}
&
\textbf{PBR. $\uparrow$}
&
\textbf{IAR. $\downarrow$}
\\

\midrule

Gemma-3-12b-it \cite{Kamath2025Gemma3T}
&9.71\%&5.71\%&82.86\%
&30.29\%&16.06\%&50.36\%
&31.61\%&25.81\%&54.84\%
&27.57\%&19.42\%&55.34\%
\\

Qwen-2.5-VL-7B-Instruct \cite{bai2025qwen25vl}
&19.71\%&14.29\%&62.86\%
&33.43\%&17.52\%&40.15\%
&21.94\%&12.90\%&61.29\%
&26.02\%&18.45\%&57.28\%
\\

Qwen-2.5-VL-32B-Instruct \cite{bai2025qwen25vl}
&44.00\%&42.86\%&51.43\%
&\colorbox{second}{47.08\%}&31.39\%&27.01\%
&37.42\%&35.48\%&45.16\%
&36.60\%&24.27\%&40.78\%
\\

Qwen-3-VL-8B-Instruct \cite{bai2025qwen3}
&8.57\%&8.57\%&91.43\%
&14.45\%&10.22\%&77.37\%
&25.16\%&29.03\%&70.97\%
&36.99\%&30.10\%&45.63\%
\\

Qwen-3-VL-32B-Instruct \cite{bai2025qwen3}
&12.86\%&11.43\%&85.71\%
&14.53\%&13.14\%&78.83\%
&13.55\%&12.90\%&83.87\%
&32.04\%&23.30\%&50.49\%
\\

InternVL3-8B-Instruct \cite{zhu2025internvl3}
&19.14\%&22.86\%&74.29\%
&12.19\%&6.57\%&78.83\%
&19.68\%&16.13\%&70.97\%
&14.17\%&10.68\%&77.67\%
\\

InternVL3-14B-Instruct \cite{zhu2025internvl3}
&19.43\%&14.29\%&68.57\%
&12.70\%&8.03\%&79.56\%
&29.35\%&32.26\%&58.06\%
&15.92\%&15.53\%&75.73\%
\\

InternVL3.5-8B \cite{wang2025internvl3}
&20.57\%&20.00\%&68.57\%
&11.82\%&7.30\%&80.29\%
&28.39\%&32.26\%&58.06\%
&20.49\%&15.53\%&66.02\%
\\

InternVL3.5-14B \cite{wang2025internvl3}
&11.14\%&8.57\%&82.86\%
&16.42\%&8.03\%&72.99\%
&24.52\%&19.35\%&64.52\%
&26.89\%&15.53\%&55.34\%
\\

\midrule

LLaVA-Med-v1.5 \cite{li2023llava}
&41.71\%&42.86\%&54.29\%
&26.64\%&21.17\%&64.23\%
&30.97\%&32.26\%&64.52\%
&27.77\%&24.27\%&64.08\%
\\

HuatuoGPT-Vision-7B \cite{chen2024towardsHuatuoGPT}
&24.00\%&28.57\%&68.57\%
&28.54\%&13.87\%&50.36\%
&21.94\%&29.03\%&67.74\%
&23.50\%&23.30\%&63.11\%
\\

Med-Flamingo-9B \cite{moor2023med}
&42.29\%&42.86\%&34.29\%
&21.17\%&17.52\%&66.42\%
&39.35\%&38.71\%&35.48\%
&29.42\%&20.39\%&50.49\%
\\

Lingshu-I-8B \cite{xu2025lingshu}
&58.29\%&68.57\%&31.43\%
&44.23\%&31.39\%&32.12\%
&49.03\%&58.06\%&35.48\%
&36.41\%&31.07\%&46.60\%
\\

MedGemma-4b-it \cite{sellergren2025medgemma}
&49.43\%&51.43\%&37.14\%
&43.80\%&24.09\%&27.74\%
&28.71\%&29.03\%&58.06\%
&31.36\%&19.42\%&48.54\%
\\

MedGemma-27b-it \cite{sellergren2025medgemma}
&23.43\%&20.00\%&62.86\%
&38.98\%&20.44\%&32.85\%
&19.03\%&22.58\%&70.97\%
&35.92\%&23.30\%&39.81\%
\\

HealthGPT-Pro-4B \cite{lin2025healthgptmedicallargevisionlanguage}
&48.29\%&48.57\%&48.57\%
&39.56\%&25.55\%&40.88\%
&34.52\%&32.26\%&61.29\%
&31.55\%&24.27\%&55.34\%
\\

HealthGPT-Pro-8B \cite{lin2025healthgptmedicallargevisionlanguage}
&\colorbox{second}{73.43\%}&74.29\%&25.71\%
&46.13\%&32.85\%&35.77\%
&52.26\%&51.61\%&45.16\%
&36.02\%&29.13\%&52.43\%
\\

\midrule

Hulu-Med-7B \cite{jiang2025hulu}
&38.86\%&45.71\%&54.29\%
&36.64\%&29.93\%&45.26\%
&27.10\%&25.81\%&67.74\%
&34.17\%&23.30\%&49.51\%
\\

Hulu-Med-14B \cite{jiang2025hulu}
&31.14\%&31.43\%&65.71\%
&39.12\%&29.20\%&40.88\%
&31.94\%&32.26\%&64.52\%
&32.14\%&24.27\%&53.40\%
\\

PathGen-LLaVA \cite{sun2025pathgen}
&39.14\%&48.57\%&51.43\%
&22.04\%&16.79\%&65.69\%
&38.71\%&45.16\%&48.39\%
&23.69\%&19.42\%&65.05\%
\\

Patho-R1-7B \cite{zhang2026patho}
&66.57\%&\colorbox{second}{82.86\%}&\colorbox{first}{8.57\%}
&36.57\%&29.20\%&41.61\%
&60.00\%&\colorbox{second}{80.65\%}&\colorbox{second}{16.13\%}
&33.20\%&\colorbox{second}{35.92\%}&51.46\%
\\

Quilt-LLaVA-v1.5-7b \cite{seyfioglu2024quilt}
&49.43\%&51.43\%&28.57\%
&38.39\%&19.71\%&33.58\%
&38.39\%&41.94\%&41.94\%
&27.96\%&17.48\%&50.49\%
\\

\midrule

WSI-VQA \cite{chen2024wsi}
&66.86\%&65.71\%&28.57\%
&20.51\%&15.33\%&72.99\%
&63.55\%&61.29\%&32.26\%
&26.31\%&20.39\%&66.02\%
\\

PA-LLaVA \cite{dai2024pa}
&72.57\%&71.43\%&22.86\%
&22.70\%&18.25\%&70.80\%
&\colorbox{second}{66.77\%}&61.29\%&25.81\%
&27.77\%&22.33\%&65.05\%
\\

WSI-LLaVA \cite{liang2025wsi}
&62.57\%&77.14\%&20.00\%
&{46.28\%}&\colorbox{second}{37.23\%}&27.74\%
&45.81\%&54.84\%&41.94\%
&38.45\%&25.24\%&35.92\%
\\

MedGemma-1.5-4b-it \cite{sellergren2026medgemma}
&49.43\%&51.43\%&28.57\%
&38.39\%&19.71\%&33.58\%
&44.84\%&54.84\%&41.94\%
&36.80\%&24.27\%&38.83\%
\\

SlideChat \cite{chen2025slidechat}
&34.57\%&34.29\%&60.00\%
&28.47\%&18.98\%&{55.47\%}
&32.58\%&32.26\%&61.29\%
&33.50\%&24.27\%&51.46\%
\\

ANTONI-Alpha \cite{moonemans2025democratizing}
&54.29\%&57.14\%&22.86\%
&43.14\%&26.28\%&\colorbox{second}{25.55\%}
&43.55\%&51.61\%&45.16\%
&37.38\%&24.27\%&37.86\%
\\

\midrule

PathAgent \cite{chen2025pathagent}
&66.86\%&65.71\%&20.00\%
&41.61\%&25.55\%&35.77\%
&55.16\%&61.29\%&25.81\%
&43.88\%&32.04\%&\colorbox{second}{29.13\%}
\\

PathNavigate \cite{yang2026pathnavigate}
&65.71\%&68.57\%&20.00\%
&43.58\%&26.28\%&29.93\%
&57.74\%&64.52\%&25.81\%
&\colorbox{second}{44.95\%}&33.98\%&30.10\%
\\

\textbf{Ours}
&\colorbox{first}{\textbf{76.00\%}}
&\colorbox{first}{\textbf{85.71\%}}
&\colorbox{second}{\textbf{11.43\%}}
&\colorbox{first}{\textbf{53.72\%}}
&\colorbox{first}{\textbf{42.34\%}}
&\colorbox{first}{\textbf{20.44\%}}
&\colorbox{first}{\textbf{69.68\%}}
&\colorbox{first}{\textbf{83.87\%}}
&\colorbox{first}{\textbf{9.68\%}}
&\colorbox{first}{\textbf{51.75\%}}
&\colorbox{first}{\textbf{51.46\%}}
&\colorbox{first}{\textbf{22.33\%}}
\\

\bottomrule[1.5pt]

\end{tabular}
\end{lrbox}

\resizebox{0.9\textwidth}{!}{\usebox{\tablebox}}
\end{table*}

\begin{figure*}[!htb]
  \centering
\includegraphics[width=0.88\textwidth]{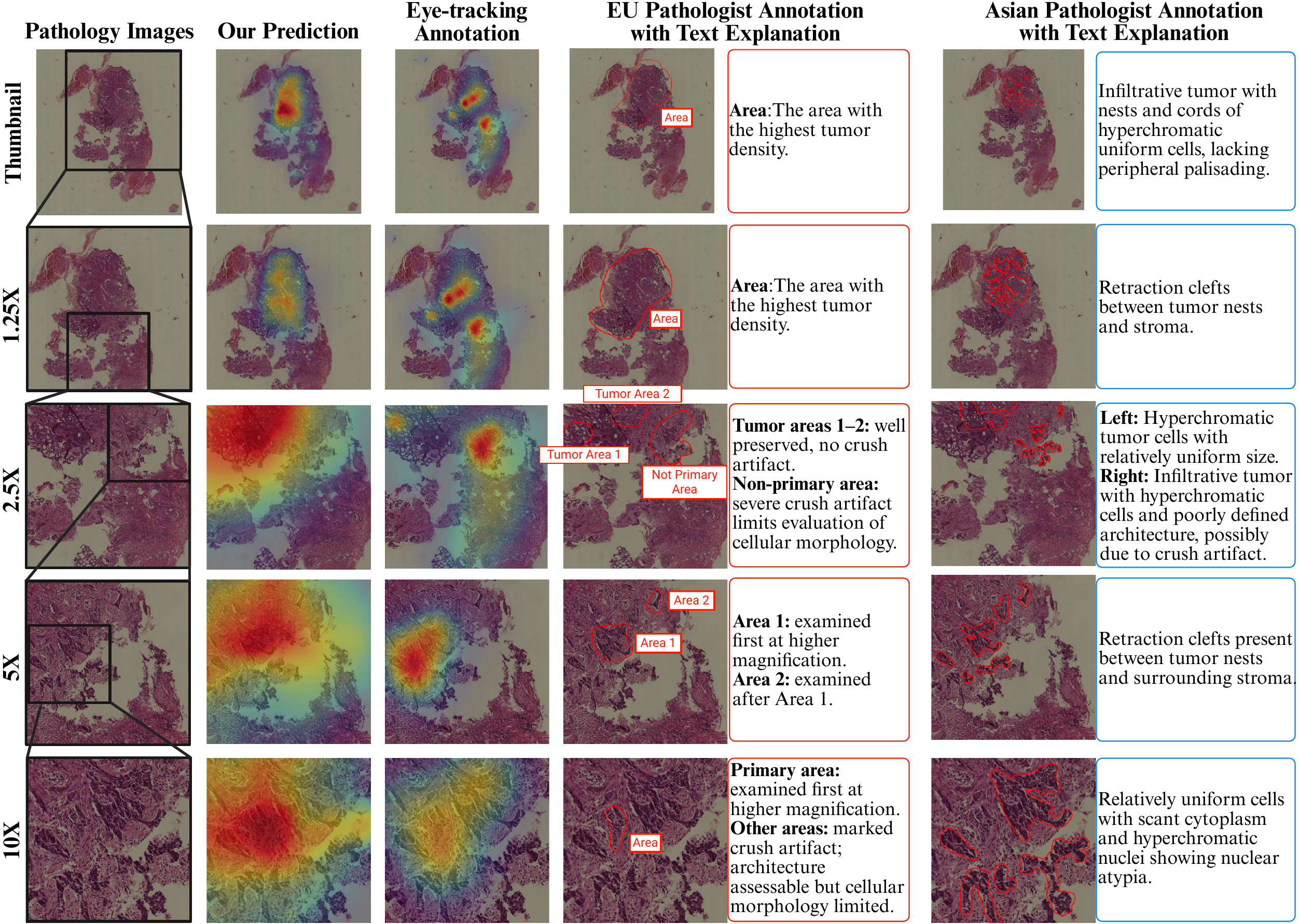}
\caption{
Pathologist-in-the-loop interpretation of CrossMagLoc across five magnifications (thumbnail, 1.25$\times$, 2.5$\times$, 5$\times$, and 10$\times$; top to bottom). The columns show the pathology images (column 1), CrossMagLoc predictions (column 2), test-set navigation annotations (column 3), and independent annotations with morphological explanations from pathologists trained in Europe (column 4) and Asia (column 5). Across magnifications, CrossMagLoc consistently prioritizes diagnostically relevant regions identified by the pathologists while deprioritizing non-primary or artifact-limited areas. At 2.5$\times$, it assigns higher attention to well-preserved tumor regions than to the crush-artifact region, while at higher magnifications it continues to emphasize morphologically informative tumor regions. Overall, its localization broadly agrees with pathologist viewing behavior while additionally highlighting clinically relevant regions identified during independent review.
} \label{fig5}
\end{figure*}

\textbf{Architectural-level QA performance.}
Architectural questions require reasoning about how pathological structures are organized and spatially related across tissue regions. Reliable answering therefore requires broader architectural context to be preserved while local interfaces, boundaries, and structural relationships are examined at sufficient detail. Unlike tissue-level QA, which primarily tests the integration of spatially distributed evidence, architectural-level QA more directly evaluates whether structural correspondence and continuity can be maintained and resolved \textbf{across regions and magnifications}.

As shown in Fig.~\ref{fig:performance} and Tables~\ref{results:tum-uteria} and \ref{results:tum-liver}, PathPilot achieves the \textbf{highest architectural GACC and PBR} across both internal cohorts.
The comparison across WSI analysis paradigms further reflects the specific challenge of architectural QA. Among thumbnail-based pathology VLMs, Patho-R1-7B achieves 60.00\% architectural GACC on TUM-Uteria and PathGen-LLaVA achieves 56.00\% on TUM Liver; although a thumbnail preserves gross tissue organization, downsampling can obscure the continuity of invasive fronts, interfaces, and fine structural boundaries. Among patch-based WSI VLMs, WSI-LLaVA achieves 56.07\% on TUM-Uteria and SlideChat achieves 64.00\% on TUM Liver; their higher-resolution patches preserve local morphology, but spatially separated patch representations can fragment the architectural relationship itself, making it difficult to determine whether structures remain continuous, cross a boundary, or extend into adjacent tissue. Among existing pathology agents, PathAgent achieves 48.93\% on TUM-Uteria and PathNavigate achieves 63.00\% on TUM Liver; although these methods can inspect selected regions, their navigation does not explicitly preserve and refine the same architectural relationship across magnifications and neighboring regions. In comparison, PathPilot achieves 65.00\% and 69.00\% on the two cohorts. CrossMagLoc provides spatially corresponding candidate ROIs across magnifications, while EviRegionSearch can follow an unresolved structure toward a finer boundary, a neighboring region, or a broader contextual view. This enables PathPilot to maintain and progressively resolve architectural relationships rather than treating each observed region as an independent piece of evidence.

\textbf{Cellular-level QA performance.}
Cellular-level questions require localized fine-grained evidence from specific cell populations or cytological structures. Because such findings are often focal or sparse, reliable answering requires the relevant cellular region to be \textbf{localized within the appropriate tissue context} and examined at sufficient visual detail. This category therefore more directly evaluates whether PathPilot can progressively refine its exploration toward \textbf{question-relevant high-resolution evidence} rather than relying on an arbitrary local field.

As shown in Fig.~\ref{fig:performance} and Tables~\ref{results:tum-uteria} and \ref{results:tum-liver}, PathPilot achieves the \textbf{highest cellular-level GACC and PBR and the lowest IAR} across both internal cohorts.
The comparison across WSI analysis paradigms further reflects the specific challenge of cellular QA. Among thumbnail-based pathology VLMs, PathGen-LLaVA achieves the strongest cellular GACC of 56.88\% on TUM-Uteria, while Patho-R1-7B achieves 53.33\% on TUM Liver; however, a single downsampled slide view may not preserve the visual detail needed to identify sparse cells or subtle cytological alterations. Among patch-based WSI VLMs, WSI-LLaVA achieves 56.67\% and 47.08\% on TUM-Uteria and TUM Liver, respectively; although higher-resolution local morphology is retained, predefined patch sampling does not ensure that the selected regions contain the specific cellular population required by the question. Among existing pathology agents, PathNavigate achieves 48.75\% and 43.75\%, respectively; targeted exploration improves access to local evidence, but constrained region and magnification refinement can still miss sparse or highly localized cellular findings. In comparison, PathPilot achieves 62.71\% and 55.83\% on the two cohorts. CrossMagLoc identifies diagnostically relevant candidate ROIs across magnifications, including high-magnification regions that may contain the cellular evidence required by the question. EviRegionSearch then progressively refines the exploration toward these candidate ROIs and redirects the search when the currently examined field remains insufficient. This is particularly important for cellular QA, where success depends not only on accessing high-resolution imagery, but on accurately localizing the relevant cellular evidence before fine-grained interpretation.

\begin{table*}[t]
\centering
\caption{Validation of the LLM-based evaluator on the morphology subsets of WSI-Bench TCGA-SKCM and TCGA-LIHC using four controlled answer variants: paraphrase (P), omission (O), attribute error (A), and core contradiction (C). The first block reports the evaluator scores for the four variants, the second block evaluates whether the expected ordering ($P \geq O \geq A \geq C$) is preserved, and the final block reports score differences between adjacent variants. Scores are reported as mean [95\% CI]. Overall, paraphrasing has little effect on the score, omission causes only a small decrease, factual attribute errors lead to a clear score reduction, and core contradictions are consistently assigned a score of 0.
}
\label{tab:judge_validation}

\setlength{\tabcolsep}{5pt}

\resizebox{0.9\textwidth}{!}{%
\begin{tabular}{llccc}
\toprule[1.5pt]
\toprule[1.5pt]

\textbf{Metric} &
\textbf{Expected} &
\textbf{WSI-Bench TCGA-SKCM} &
\textbf{WSI-Bench TCGA-LIHC} &
\textbf{Pooled} \\

\midrule
\multicolumn{5}{l}{\textit{Variant-level judge scores}} \\
\midrule

Paraphrase ($S_P$)
& High
& 0.957 [0.923, 0.991]
& 0.903 [0.855, 0.952]
& 0.932 [0.900, 0.959] \\

Omission ($S_O$)
& $\approx S_P$
& 0.829 [0.769, 0.886]
& 0.845 [0.787, 0.894]
& 0.836 [0.797, 0.876] \\

Attribute error ($S_A$)
& Lower
& 0.611 [0.560, 0.654]
& 0.348 [0.252, 0.439]
& 0.488 [0.427, 0.545] \\

Core contradiction ($S_C$)
& Lowest
& 0.000 [0.000, 0.000]
& 0.000 [0.000, 0.000]
& 0.000 [0.000, 0.000] \\

\midrule
\multicolumn{5}{l}{\textit{Overall ordered sensitivity}} \\
\midrule

Friedman test
& $p \downarrow$
& $\chi^2(3)=92.29$, $p<0.001$
& $\chi^2(3)=84.31$, $p<0.001$
& $\chi^2(3)=174.10$, $p<0.001$ \\

Kendall's $W$
& $W \uparrow$
& 0.879
& 0.907
& 0.879 \\

Monotonic consistency
& $S_P\ge S_O\ge S_A\ge S_C$
& 0.886
& 0.871
& 0.879 \\

Paraphrase preservation
& $S_P\ge0.7$
& 1.000
& 1.000
& 1.000 \\

Contradiction rejection
& $S_C\le0.3$
& 1.000
& 1.000
& 1.000 \\

\midrule
\multicolumn{5}{l}{\textit{Paired score degradation}} \\
\midrule

$P-O$
& $\Delta S\approx0$
& 0.129 [0.060, 0.197], $p_H=0.002$
& 0.058 [$-$0.010, 0.126], $p_H=0.109$
& 0.095 [0.045, 0.144], $p_H<0.001$ \\

$O-A$
& $\Delta S>0$
& 0.217 [0.151, 0.283], $p_H<0.001$
& 0.497 [0.384, 0.613], $p_H<0.001$
& 0.348 [0.279, 0.421], $p_H<0.001$ \\

$A-C$
& $\Delta S>0$
& 0.611 [0.560, 0.654], $p_H<0.001$
& 0.348 [0.252, 0.439], $p_H<0.001$
& 0.488 [0.427, 0.545], $p_H<0.001$ \\

\bottomrule[1.5pt]
\bottomrule[1.5pt]

\end{tabular}%
}

\end{table*}

\begin{table*}[t]
\centering
\setlength{\fboxsep}{0pt}
\setlength{\tabcolsep}{4.2pt}

\caption{Evaluation of the key designs of EviRegionSearch on the external WSI-Bench TCGA-SKCM dataset, including evidence requirement reasoning and evidence-to-action translation through Move, Zoom In/Zoom Out, and Stop. Results are reported for both morphology-focused questions and all QA. \colorbox{first}{Best} and \colorbox{second}{second-best} results are highlighted. \textbf{Boldface} indicates the configuration adopted in our final agent.
}
\label{results:navigation_ablation}

\begin{lrbox}{\tablebox}
\begin{tabular}{lcccccccccc}
\toprule[1.5pt]
\toprule[1.5pt]

\textbf{Candidate Prior} &
\textbf{Q-guided Region Select.} &
\textbf{Evidence Feedback} &
\textbf{Mag. Policy} &
\textbf{Stop} &
\multicolumn{3}{c}{\textbf{Morphology}} &
\multicolumn{3}{c}{\textbf{All QA (supporting)}}\\

\cmidrule(lr){6-8}
\cmidrule(lr){9-11}

&&&&&
\textbf{GACC.$\uparrow$} &
\textbf{PBR.$\uparrow$} &
\textbf{IAR.$\downarrow$} &
\textbf{GACC.$\uparrow$} &
\textbf{PBR.$\uparrow$} &
\textbf{IAR.$\downarrow$}
\\

\midrule

\multicolumn{11}{l}{\textit{evidence requirement reasoning}}\\
\midrule

\textbf{CrossMagLoc}
& \textbf{\checkmark}
& Final-only
& \textbf{Adaptive}
& \textbf{Adaptive}
& 66.29\%
& 74.29\%
& 22.86\%
& 46.50\%
& \colorbox{second}{40.15\%}
& 30.66\%
\\

\midrule
\multicolumn{11}{l}{\textit{Evidence-to-Action Translation: Move}}\\
\midrule

Random
& --
& \textbf{Progressive}
& \textbf{Adaptive}
& \textbf{Adaptive}
& 55.14\%
& 54.29\%
& 28.57\%
& 40.00\%
& 30.66\%
& 40.15\%
\\

Random
& \textbf{\checkmark}
& \textbf{Progressive}
& \textbf{Adaptive}
& \textbf{Adaptive}
& 68.00\%
& 74.29\%
& 22.86\%
& 45.18\%
& 38.69\%
& 34.31\%
\\

\textbf{CrossMagLoc}
& --
& \textbf{Progressive}
& \textbf{Adaptive}
& \textbf{Adaptive}
& \colorbox{second}{70.00\%}
& \colorbox{second}{77.14\%}
& \colorbox{second}{20.00\%}
& 47.15\%
& 39.42\%
& 30.66\%
\\

\midrule
\multicolumn{11}{l}{\textit{Evidence-to-Action Translation: Zoom In / Zoom Out}}\\
\midrule

\textbf{CrossMagLoc}
& \textbf{\checkmark}
& \textbf{Progressive}
& Fixed 1.25$\times$
& \textbf{Adaptive}
& 33.14\%
& 25.71\%
& 45.71\%
& 37.59\%
& 21.90\%
& 34.31\%
\\

\textbf{CrossMagLoc}
& \textbf{\checkmark}
& \textbf{Progressive}
& Fixed 2.5$\times$
& \textbf{Adaptive}
& 45.43\%
& 40.00\%
& 37.14\%
& 41.09\%
& 25.55\%
& 32.12\%
\\

\textbf{CrossMagLoc}
& \textbf{\checkmark}
& \textbf{Progressive}
& Fixed 5$\times$
& \textbf{Adaptive}
& 46.86\%
& 42.86\%
& 28.57\%
& 44.16\%
& 29.20\%
& 24.09\%
\\

\textbf{CrossMagLoc}
& \textbf{\checkmark}
& \textbf{Progressive}
& Fixed 10$\times$
& \textbf{Adaptive}
& 47.14\%
& 37.14\%
& 22.86\%
& 43.43\%
& 26.28\%
& \colorbox{second}{22.63\%}
\\

\textbf{CrossMagLoc}
& \textbf{\checkmark}
& \textbf{Progressive}
& Coarse-to-fine
& \textbf{Adaptive}
& 65.71\%
& 71.43\%
& 22.86\%
& \colorbox{second}{47.37\%}
& 33.58\%
& 27.01\%
\\

\midrule
\multicolumn{11}{l}{\textit{Evidence-to-Action Translation: Stop}}\\
\midrule

\textbf{CrossMagLoc}
& \textbf{\checkmark}
& \textbf{Progressive}
& \textbf{Adaptive}
& Fixed budget
& 49.71\%
& 48.57\%
& 37.14\%
& 44.45\%
& 32.85\%
& 29.93\%
\\

\midrule

\textbf{CrossMagLoc}
& \textbf{\checkmark}
& \textbf{Progressive}
& \textbf{Adaptive}
& \textbf{Adaptive}
&\colorbox{first}{\textbf{76.00\%}}
&\colorbox{first}{\textbf{85.71\%}}
&\colorbox{first}{\textbf{11.43\%}}
& \textbf{\colorbox{first}{53.72\%}}
& \textbf{\colorbox{first}{42.34\%}}
& \textbf{\colorbox{first}{20.44\%}}
\\

\bottomrule[1.5pt]
\bottomrule[1.5pt]

\end{tabular}
\end{lrbox}

\resizebox{0.9\textwidth}{!}{\usebox{\tablebox}}
\end{table*}

\begin{table*}[t]
\centering
\setlength{\fboxsep}{0pt}

\caption{
Ablation study of PathPilot under different LLM, VLM, and visual prompting configurations on the external WSI-Bench TCGA-SKCM dataset.
\colorbox{first}{Best} and \colorbox{second}{second-best} results are highlighted.
\textbf{Boldface} indicates the configuration adopted in our final agent.
}
\label{results:backbone_ablation}

\begin{lrbox}{\tablebox}
\begin{tabular}{lllcccccc}
\toprule[1.5pt]
\toprule[1.5pt]
\multicolumn{1}{c}{\textbf{LLM}} &
\multicolumn{1}{c}{\textbf{VLM}} &
\multicolumn{1}{c}{\textbf{VLM Prompt}} &
\multicolumn{3}{c}{
\shortstack{\textbf{Morphology}\\[-0.45ex]\rule{3.25cm}{0.9pt}}
} &
\multicolumn{3}{c}{
\shortstack{\textbf{All QA (supporting)}\\[-0.45ex]\rule{3.25cm}{0.9pt}}
}
\\[-0.35ex]

&&&
\textbf{GACC. $\uparrow$} &
\textbf{PBR. $\uparrow$} &
\textbf{IAR. $\downarrow$} &
\textbf{GACC. $\uparrow$} &
\textbf{PBR. $\uparrow$} &
\textbf{IAR. $\downarrow$}
\\

\midrule

\multicolumn{9}{l}{\textit{LLM Backbone}}\\
\midrule

Mistral-Small-3.2-24B-Instruct-2506 \cite{mistralsmall32} &
\textbf{Patho-R1-7B} &
\textbf{Visual guidance} &
58.57\% & 68.57\% & 25.71\% &
40.58\% & 35.77\% & 39.42\%
\\

Phi-4 (14B) \cite{abdin2024phi} &
\textbf{Patho-R1-7B} &
\textbf{Visual guidance} &
58.29\% & 62.86\% & 34.29\% &
46.93\% & 30.66\% & 28.47\%
\\

DeepSeek-R1-Distill-Qwen-14B \cite{guo2025deepseek} &
\textbf{Patho-R1-7B} &
\textbf{Visual guidance} &
63.14\% & 65.71\% & 25.71\% &
47.45\% & 37.96\% & 29.93\%
\\

Qwen2.5-7B &
\textbf{Patho-R1-7B} &
\textbf{Visual guidance} &
67.43\% &
\colorbox{second}{77.14\%} &
20.00\% &
34.38\% & 29.93\% & 51.09\%
\\

Qwen2.5-14B &
\textbf{Patho-R1-7B} &
\textbf{Visual guidance} &
65.14\% & 62.86\% & 25.71\% &
\colorbox{second}{50.66\%} &
35.04\% &
\colorbox{second}{24.09\%}
\\

Qwen2.5-32B &
\textbf{Patho-R1-7B} &
\textbf{Visual guidance} &
\colorbox{second}{69.71\%} &
68.57\% &
\colorbox{second}{17.14\%} &
48.91\% &
32.85\% &
\colorbox{second}{24.09\%}
\\

Qwen3-8B &
\textbf{Patho-R1-7B} &
\textbf{Visual guidance} &
67.14\% & 71.43\% & 20.00\% &
45.62\% & 34.31\% & 32.12\%
\\

Qwen3-32B &
\textbf{Patho-R1-7B} &
\textbf{Visual guidance} &
67.14\% & 74.29\% & 22.86\% &
48.83\% &
\colorbox{second}{40.88\%} &
27.01\%
\\

\textbf{Qwen3-14B} &
\textbf{Patho-R1-7B} &
\textbf{Visual guidance} &
\colorbox{first}{\textbf{76.00\%}}
&\colorbox{first}{\textbf{85.71\%}}
&\colorbox{first}{\textbf{11.43\%}}&
\colorbox{first}{53.72\%} &
\colorbox{first}{42.34\%} &
\colorbox{first}{20.44\%}
\\

\midrule
\multicolumn{9}{l}{\textit{VLM Backbone}}\\
\midrule

\textbf{Qwen3-14B} &
HealthGPT-Pro-8B &
\textbf{Visual guidance} &
\colorbox{second}{75.14\%} &
\colorbox{second}{77.14\%} &
20.00\% &
\colorbox{second}{46.13\%} &
32.85\% &
\colorbox{second}{35.77\%}
\\

\textbf{Qwen3-14B} &
Quilt-LLaVA-v1.5-7b &
\textbf{Visual guidance} &
72.86\% &
\colorbox{second}{77.14\%} &
\colorbox{second}{14.29\%} &
42.48\% &
33.58\% &
37.23\%
\\

\textbf{Qwen3-14B} &
MedGemma-1.5-4b-it &
\textbf{Visual guidance} &
68.86\% & 71.43\% & 20.00\% &
44.16\% &
\colorbox{second}{39.42\%} &
\colorbox{second}{35.77\%}
\\

\textbf{Qwen3-14B} &
\textbf{Patho-R1-7B} &
\textbf{Visual guidance} &
\colorbox{first}{\textbf{76.00\%}}
&\colorbox{first}{\textbf{85.71\%}}
&\colorbox{first}{\textbf{11.43\%}}&
\colorbox{first}{53.72\%} &
\colorbox{first}{42.34\%} &
\colorbox{first}{20.44\%}
\\

\midrule
\multicolumn{9}{l}{\textit{Visual Prompting}}\\
\midrule

\textbf{Qwen3-14B} &
\textbf{Patho-R1-7B} &
Without Question &
62.86\% &
71.43\% &
{20.00\%} &
41.39\% &
35.77\% &
37.23\%
\\

\textbf{Qwen3-14B} &
\textbf{Patho-R1-7B} &
Original Question &
\colorbox{second}{74.86\%} &
\colorbox{second}{82.86\%} &
\colorbox{second}{14.29\%} &
\colorbox{second}{45.77\%} &
\colorbox{second}{40.88\%} &
\colorbox{second}{35.04\%}
\\

\textbf{Qwen3-14B} &
\textbf{Patho-R1-7B} &
\textbf{Visual guidance} &
\colorbox{first}{\textbf{76.00\%}}
&\colorbox{first}{\textbf{85.71\%}}
&\colorbox{first}{\textbf{11.43\%}}&
\colorbox{first}{53.72\%} &
\colorbox{first}{42.34\%} &
\colorbox{first}{20.44\%}
\\

\bottomrule[1.5pt]
\bottomrule[1.5pt]

\end{tabular}
\end{lrbox}

\scalebox{0.73}{\usebox{\tablebox}}
\end{table*}

\subsubsection{External Validation.}

We further evaluate PathPilot on three external cohorts, including TUM Skin, TCGA-SKCM, and TCGA-LIHC, to assess the generalizability of PathPilot. TUM Skin provides an external clinical setting in which performance can be examined across tissue-level, architectural, and cellular findings, while the two WSI-Bench cohorts introduce independently constructed question distributions and provide morphology-focused evaluation that more directly depends on localized, multi-magnification H\&E evidence. Together, these settings allow us to assess both the \textbf{broader robustness} of PathPilot and the \textbf{transferability} of its visual evidence acquisition strategy.

\textbf{Average QA performance.}
As shown in Fig.~\ref{fig:performance} and Tables~\ref{results:tum-skin}--\ref{results:tcga_external}, PathPilot consistently achieves the \textbf{highest average GACC} across all three external cohorts, reaching 61.59\% on TUM Skin, 53.72\% on TCGA-SKCM, and 51.75\% on TCGA-LIHC. On TUM Skin, the GACC remains essentially unchanged from the two internal cohorts (60.36\% and 60.47\%), while exceeding the strongest external baseline by 13.64 percentage points. On TCGA-SKCM and TCGA-LIHC, where the question distributions differ substantially from the TUM cohorts, PathPilot still achieves the highest average GACC, outperforming the strongest baselines by 6.64 and 6.80 percentage points, respectively. This consistent lead demonstrates that the performance advantage observed internally is preserved across external cohorts with different pathological and question distributions. Because the full WSI-Bench benchmark also contains question types that are not primarily determined by localized H\&E morphology, we treat these all-QA results as supporting evidence of \textbf{broader robustness}, while using the morphology subsets below as a more focused evaluation of visual evidence acquisition.

\textbf{Tissue-level QA performance.}
PathPilot achieves the \textbf{highest tissue-level GACC} of 65.21\% on TUM Skin, compared with 57.69\% and 63.33\% on TUM-Uteria and TUM Liver, respectively. The external result therefore shows no degradation relative to the internal cohorts and exceeds the strongest TUM Skin baseline by 10.69 percentage points. This indicates robust tissue-level performance under external cohort shift, preserving the ability to answer questions requiring broader tissue-level evidence.

\textbf{Architectural-level QA performance.}
For architectural questions, PathPilot achieves a GACC of 53.75\% on TUM Skin, lower than the 65.00\% and 69.00\% obtained on the two internal cohorts. Nevertheless, it \textbf{remains the best-performing method} on the external cohort and exceeds the strongest baseline by 8.75 percentage points. Thus, despite the lower absolute performance on external architectural questions, PathPilot maintains a clear relative advantage in resolving spatial and structural relationships.

\textbf{Cellular-level QA performance.}
For cellular questions, PathPilot achieves the \textbf{highest GACC} of 61.11\% on TUM Skin, remaining within the range observed on the internal cohorts (62.71\% on TUM-Uteria and 55.83\% on TUM Liver). It also exceeds the strongest external baseline by 9.63 percentage points, while maintaining a low IAR of 3.70\%. This combination of leading GACC and a low error rate supports robust external performance for questions requiring localized fine-grained evidence, preserving both the absolute performance level and the relative advantage observed internally.

\textbf{Morphology-level QA performance.}
Morphology-level performance provides the most direct external assessment of the proposed navigation capability, since these questions primarily depend on retrieving and integrating H\&E-visible morphological evidence from appropriate regions and magnification levels. As shown in Table~\ref{results:tcga_external}, PathPilot achieves morphology GACC values of 76.00\% on TCGA-SKCM and 69.68\% on TCGA-LIHC, exceeding the strongest baselines by 2.56 and 2.91 percentage points, respectively, with corresponding PBR values of 85.71\% and 83.87\%. It achieves the \textbf{highest GACC and PBR} on TCGA-SKCM and the \textbf{best performance} across all three metrics on TCGA-LIHC.

This consistent advantage across both TCGA cohorts supports the generalizability of evidence acquisition under changes in cohort morphology and question requirements. CrossMagLoc provides transferable, pathologist-derived diagnostic spatial guidance across magnifications, while EviRegionSearch updates the accumulated question-relevant evidence and redirects exploration when previously examined regions do not sufficiently resolve the current question.

\subsubsection{Validation of the LLM-Based Evaluator.}
\label{sec:judge_validation}

To assess whether the evaluator reflects answer correctness, we construct controlled answer perturbations on the morphology subsets of TCGA-SKCM and TCGA-LIHC, comprising 35 and 31 QA groups, respectively. Each reference answer yields four variants: (1) \textit{paraphrase}, preserving meaning while changing wording; (2) \textit{omission}, removing one supporting finding while retaining the core answer; (3) \textit{attribute error}, introducing an incorrect morphological, quantitative, spatial, or extent-related attribute while preserving the remaining content; and (4) \textit{core contradiction}, reversing or negating the principal finding. To reduce self-evaluation bias, GPT-5.6 Sol \cite{openai2026gpt56} generates these variants independently of the Qwen-based evaluator, and pathologists verify the intended semantic changes. The finalized perturbation set will be released with our code.

We report mean evaluator scores with bootstrap 95\% confidence intervals and use paired comparisons because all four variants originate from the same QA. Overall differences are assessed using the Friedman test with Kendall's $W$ as the effect size; adjacent levels are compared using paired Wilcoxon signed-rank tests with Holm correction. We also measure the proportion of QA groups satisfying the expected ordering $S_P \geq S_O \geq S_A \geq S_C$, where $P$, $O$, $A$, and $C$ denote the four variants above. Paraphrase preservation and contradiction rejection are defined as the proportions of paraphrases scoring at least 0.7 and core contradictions scoring at most 0.3, respectively.

As shown in Table~\ref{tab:judge_validation}, scores decrease with increasing error severity on both datasets. Mean scores for paraphrase, omission, attribute error, and core contradiction are 0.957, 0.829, 0.611, and 0.000 on TCGA-SKCM, and 0.903, 0.845, 0.348, and 0.000 on TCGA-LIHC. Overall differences are significant with large effect sizes (TCGA-SKCM: $\chi^2(3)=92.29$, $W=0.879$; TCGA-LIHC: $\chi^2(3)=84.31$, $W=0.907$; both $p<0.001$). The expected ordering holds in 0.886 and 0.871 of QA groups, respectively, or 0.879 overall. Pooled scores decrease from 0.932 to 0.836, 0.488, and 0.000: omission produces a relatively small reduction ($\Delta S=0.095$), whereas attribute errors and core contradictions produce larger successive reductions ($\Delta S=0.348$ and 0.488; both $p_H<0.001$). Paraphrase preservation and contradiction rejection both reach 1.000 on each dataset. Together, these results demonstrate the evaluator's \textbf{robustness to paraphrasing} and \textbf{sensitivity to factual errors}, supporting the \textbf{reliability of the reported QA evaluations}.

\begin{figure}[!t]
  \centering
\includegraphics[width=1.0\columnwidth]{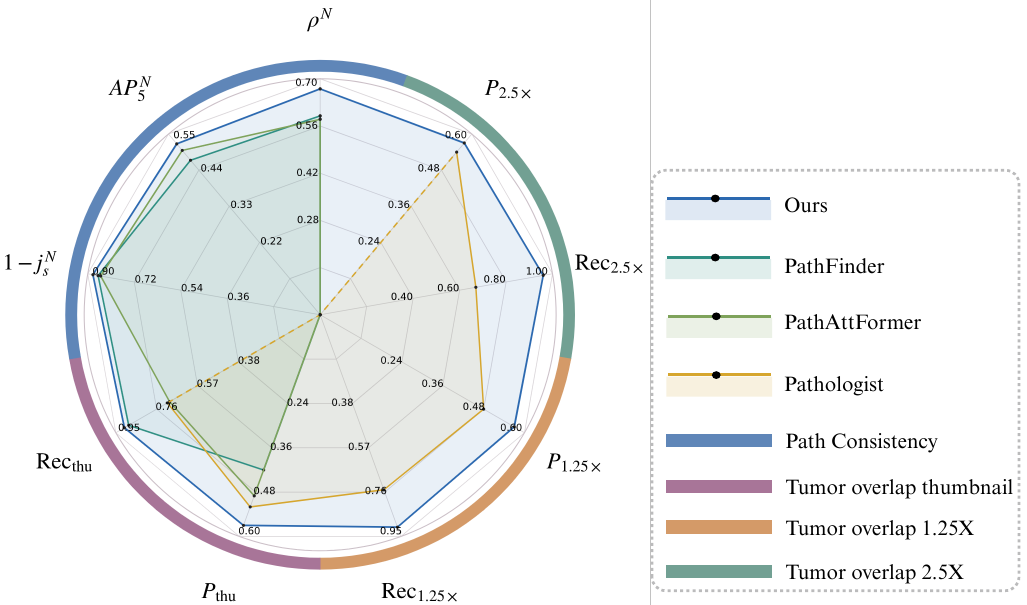}
\caption{\textbf{Quantitative evaluation of CrossMagLoc navigation performance.} The radar plot summarizes pathologist-prioritization alignment ($\rho^N$, $AP_5^N$, and $1-JS^N$) and tumor-region precision/recall at the thumbnail, 1.25$\times$, and 2.5$\times$ magnifications, with PathFinder \cite{ghezloo2025pathfinder} and PathAttFormer \cite{chakraborty2022predicting} as navigation baselines and pathologist navigation as a clinical reference for tumor-region alignment. CrossMagLoc achieves the strongest agreement with pathologist prioritization while maintaining high tumor recall across magnifications, indicating a similar balance between diagnostic focus and broad tumor coverage.
} \label{fig:cmt_performance}
\end{figure}

\subsection{Quantitative Navigation Performance of CrossMagLoc}
\label{sec:nav_exp}

As shown in Fig.~\ref{fig:cmt_performance}, we further validate whether CrossMagLoc localizes diagnostically relevant regions through region-level alignment analysis on the Skin Eye-Tracking dataset, which provides both pathologist eye-tracking annotations and pixel-level tumor masks. We evaluate the predicted navigation heatmaps from two complementary perspectives: \textit{pathologist-attention alignment}, measuring \textbf{consistency with expert region prioritization}, and \textit{tumor-region alignment}, measuring the \textbf{relationship between high-attention regions and annotated tumor tissue}. CrossMagLoc is compared with PathFinder \cite{ghezloo2025pathfinder} and PathAttFormer \cite{chakraborty2022predicting}, two representative thumbnail-based navigation methods. Unlike these baselines, CrossMagLoc generates magnification-specific heatmaps at the thumbnail, 1.25$\times$, 2.5$\times$, 5$\times$, and 10$\times$ levels. Pathologist navigation annotations from the test set are additionally evaluated under the tumor-region metrics as a clinical reference.

\subsubsection{Alignment with Pathologist Region Prioritization.}
We first assess whether the regions prioritized by CrossMagLoc are consistent with those prioritized by pathologists during WSI examination. Predicted heatmaps are compared with pathologist viewing heatmaps using three complementary metrics: Spearman rank correlation $\rho^{N}$  \cite{spearman1961proof} for agreement in regional importance ranking, $AP^{N}_{5}$ for retrieval of the top 5\% of regions ranked by pathologist, and Jensen--Shannon divergence $JS^{N}$  \cite{lin2002divergence}. We report $1-JS^{N}$ so that higher values consistently indicate better alignment. Together, these metrics characterize regional ranking, expert-attended hotspot retrieval, and global attention-distribution similarity. Since the pathologist annotations themselves serve as the reference, their self-alignment would be trivially perfect and is therefore not reported as a separate baseline.

CrossMagLoc achieves the \textbf{strongest alignment with pathologist attention}, with $\rho^{N}=0.70$, $AP^{N}_{5}=0.55$, and $1-JS^{N}=0.90$. These results indicate that the predicted heatmaps align with \textbf{pathologists' regional importance rankings} and \textbf{spatial attention distributions}. PathFinder and PathAttFormer also capture some coarse expert-attended regions, but their localization is derived from a fixed thumbnail representation. In contrast, CrossMagLoc preserves pathologist-guided region prioritization across multiple magnifications, providing a \textbf{cross-scale diagnostic-region prior} for subsequent question-conditioned evidence acquisition rather than requiring exact reproduction of the expert attention map.

\begin{figure*}[h]
  \centering
\includegraphics[width=1.0\textwidth]{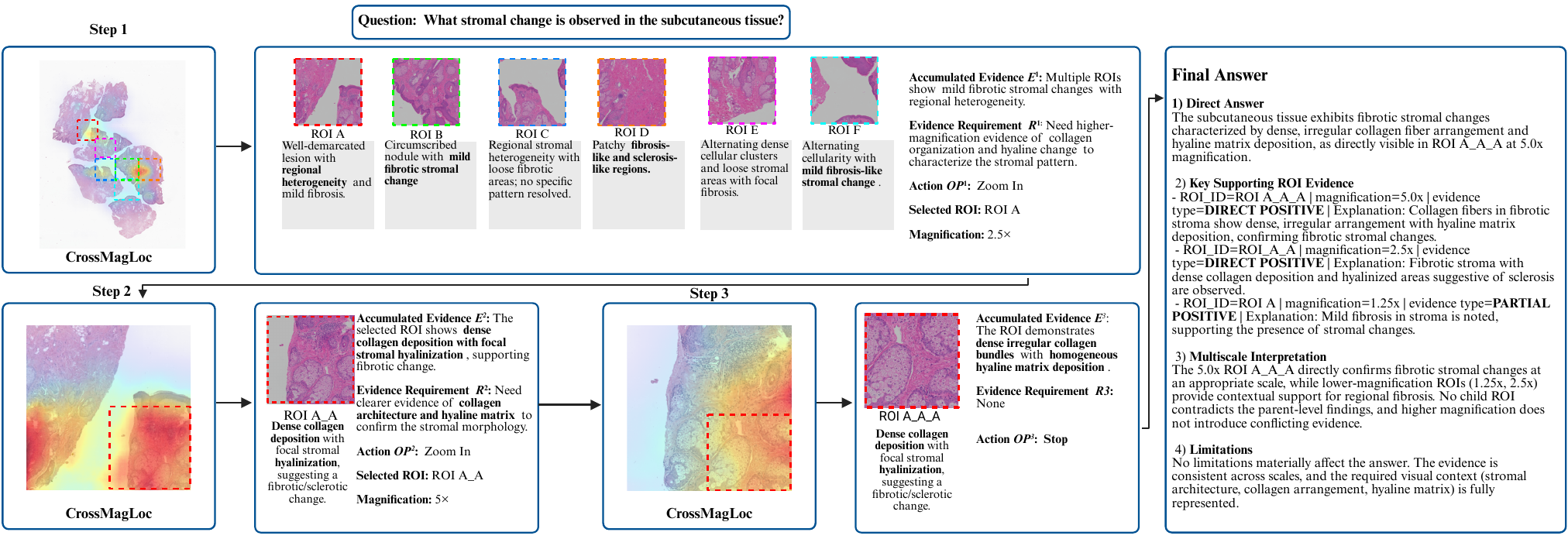}
\caption{\textbf{Visualization of a representative question-conditioned navigation workflow.} CrossMagLoc-based ROI localization and EviRegionSearch-based evidence reasoning are iteratively coupled to support adaptive exploration across regions and magnifications, evidence requirement reasoning, and final answer generation once sufficient evidence is acquired.}\label{fig6}
\end{figure*}

\subsubsection{Alignment with Tumor Regions.}

We next evaluate whether the high-attention regions identified by CrossMagLoc primarily focus on tumor tissue while also extending beyond tumor boundaries to potentially informative tissue context. Ranked tumor precision $P$ measures the proportion of tumor tissue within the top 10\% highest-scoring navigation regions, while tumor recall $Rec$ measures the proportion of annotated tumor area covered by these regions. Together, these metrics characterize tumor focus and coverage, while allowing diagnostically relevant attention to extend to non-tumor regions.

Tumor-region alignment is evaluated at the thumbnail, 1.25$\times$, and 2.5$\times$ levels. Although CrossMagLoc also generates heatmaps at 5$\times$ and 10$\times$, the available masks delineate gross tumor extent without identifying diagnostically relevant tissue structures or cellular morphology within these regions. We therefore restrict this analysis to lower magnifications, where tumor extent provides a meaningful spatial reference. PathFinder and PathAttFormer are evaluated only at the thumbnail level because their heatmaps are generated from a fixed low-resolution WSI representation.

CrossMagLoc achieves ranked tumor precision values of 0.60, 0.80, and 0.60 at the thumbnail, 1.25$\times$, and 2.5$\times$ levels, respectively, with corresponding tumor recall values of 0.85, 0.95, and 1.00. These results show that the prioritized regions are \textbf{predominantly tumor-focused} and maintain \textbf{broad tumor coverage}, while also including tissue outside the annotated tumor regions. This pattern is consistent with diagnostic exploration that examines both tumor morphology and its tissue context; the diagnostic relevance of the additional non-tumor regions is further examined through independent pathologist review.

\subsubsection{Intended Role of CrossMagLoc.}

By jointly comparing the pathologist-attention alignment and tumor-region alignment results above, we further observe a characteristic navigation pattern. Pathologist attention is \textbf{tumor-focused but not tumor-exhaustive}: experts selectively inspect diagnostically informative tumor and contextual regions rather than covering the entire tumor mask. 

CrossMagLoc exhibits a \textbf{similar non-tumor-exclusive pattern}, with its high-attention regions remaining substantially tumor-focused while retaining broader tumor coverage for subsequent question-conditioned exploration. This suggests that CrossMagLoc captures not only the regions prioritized by pathologists, but also a similar balance between diagnostic focus and evidence coverage. This should not be interpreted as outperforming pathologists in navigation; rather, \textbf{it reflects the intended role of CrossMagLoc as a diagnostic-region prior} that preserves a sufficiently broad evidence pool for downstream reasoning.

\begin{table*}[t]
\centering
\setlength{\fboxsep}{0pt}

\caption{Ablation study of the loss terms and architectural components of CrossMagLoc using four-class MIL classification. Results are reported as mean AUC $\pm$ standard deviation across five folds at four patch budgets. \colorbox{first}{Best} and \colorbox{second}{second-best} results are highlighted.} \label{table8}
\begin{lrbox}{\tablebox}
\begin{tabular}{ccccccllll}
\toprule[1.5pt]
\toprule[1.5pt]
 \multicolumn{3}{c}{ \shortstack{\textbf{Loss}\\[-0.45ex]\rule{3.25cm}{0.9pt}}} & \multicolumn{3}{c}{ \shortstack{\textbf{Component}\\[-0.45ex]\rule{3.25cm}{0.9pt}}} & \multicolumn{4}{c}{ \shortstack{\textbf{Patch Budget}\\[-0.45ex]\rule{8.50cm}{0.9pt}}}  \\
\multicolumn{1}{c}{\makebox[1.2cm][c]{$\mathcal{L}_{1}$}} &
\multicolumn{1}{c}{\makebox[1.2cm][c]{$\mathcal{L}_{dice}$}} &
\multicolumn{1}{c}{\makebox[1.2cm][c]{$\mathcal{L}_{sig}$}} & \multicolumn{1}{c}{\textbf{IMCA}} & \multicolumn{1}{c}{\textbf{CMCF-L}} & \multicolumn{1}{c}{\textbf{CMCF-H}} & \multicolumn{1}{c}{20\%} & \multicolumn{1}{c}{40\%} & \multicolumn{1}{c}{{60\%}} & \multicolumn{1}{c}{80\%}  \\
\toprule[1pt]
 &$\checkmark$ &$\checkmark$ &$\checkmark$&$\checkmark$&$\checkmark$&90.87\%$(\pm 1.63\%)$&92.06\%$(\pm 1.01\%)$&92.66\%$(\pm 1.54\%)$&92.61\%$(\pm 0.71\%)$\\
$\checkmark$ & $\checkmark$& &$\checkmark$&$\checkmark$&$\checkmark$&91.98\%$(\pm 1.04\%)$&92.82\%$(\pm 0.99\%)$&\colorbox{second}{92.97\%}$(\pm 0.41\%)$&{92.88\%}$(\pm 0.64\%)$\\
$\checkmark$ & &$\checkmark$ &$\checkmark$&$\checkmark$&$\checkmark$&92.01\%$(\pm 0.91\%)$&\colorbox{second}{92.84\%}$(\pm 1.13\%)$&92.85\%$(\pm 0.37\%)$&92.54\%$(\pm 0.78\%)$\\
$\checkmark$ & $\checkmark$ &$\checkmark$ &&&&91.90\%$(\pm 1.50\%)$&92.83\%$(\pm 1.52\%)$&92.47\%$(\pm 0.38\%)$&92.56\%$(\pm 0.74\%)$ \\
$\checkmark$ & $\checkmark$ &$\checkmark$ &$\checkmark$&&&91.74\%$(\pm 0.37\%)$&92.65\%$(\pm 0.94\%)$&92.78\%$(\pm 0.42\%)$&92.78\%$(\pm 0.69\%)$ \\
$\checkmark$ & $\checkmark$ &$\checkmark$ &$\checkmark$&$\checkmark$&&\colorbox{second}{93.23\%}$(\pm 0.82\%)$&92.70\%$(\pm 0.45\%)$&92.91\%$(\pm 0.50\%)$& \colorbox{second}{93.34\%}$(\pm 0.78\%)$\\
$\checkmark$ & $\checkmark$ &$\checkmark$ &$\checkmark$&&$\checkmark$&93.02\%$(\pm 0.38\%)$&92.50\%$(\pm 0.91\%)$&92.92\%$(\pm 1.69\%)$&92.62\%$(\pm 0.69\%)$ \\

$\checkmark$ & $\checkmark$ &$\checkmark$ &$\checkmark$&$\checkmark$&$\checkmark$&\colorbox{first}{93.89\%}$(\pm 1.73\%)$&\colorbox{first}{93.29\%}$(\pm 1.12\%)$&\colorbox{first}{94.17}\%$(\pm 0.42\%)$& \colorbox{first}{93.90}\%$(\pm 0.73\%)$ \\

\bottomrule[1.5pt]

\end{tabular}

\end{lrbox}
\scalebox{0.76}[0.73]{\usebox{\tablebox}}
\end{table*}

\subsection{Quantitative Evidence-adaptive Exploration Performance of EviRegionSearch}
\label{sec:eviregionsearch_eval}
We next evaluate whether EviRegionSearch can adapt its multi-magnification exploration to the visual evidence (ROIs) according to the current question and remaining evidence requirement. Existing pathology navigation methods do not support such evidence-adaptive exploration across magnifications, precluding a direct component-level comparison. We therefore assess EviRegionSearch through controlled ablation studies of its two key modules, Evidence Requirement Reasoning and Evidence-to-Action Translation, varying one module at a time while keeping the other fixed.

Specifically, we first assess whether newly acquired observations need to be progressively integrated into the evolving evidence state during exploration. Second, we evaluate whether this evidence state can adaptively determine the subsequent navigation actions, including Move, Zoom In, Zoom Out, and Stop.

\subsubsection{Necessity of Evidence Requirement Reasoning.}

We first evaluate whether progressively accumulated evidence should influence subsequent navigation decisions. In the full setting, each newly acquired ROI observation is integrated into the accumulated question-relevant evidence $\mathcal{E}^{t}$ and remaining evidence requirement $\mathcal{R}^{t}$ after each navigation step. As a comparison, the final-only variant retains all ROI observations but aggregates them only before final answer generation, thereby removing evidence feedback during navigation.

As shown in Table~\ref{results:navigation_ablation}, the final-only setting reduces morphology GACC from 76.00\% to 66.29\% and PBR from 85.71\% to 74.29\%, while increasing IAR from 11.43\% to 22.86\%. Over all QA, GACC similarly decreases from 53.72\% to 46.50\%, while IAR increases from 20.44\% to 30.66\%.

These results indicate that \textbf{evidence requirement reasoning contributes beyond combining observations for final answer generation}. Maintaining an evolving evidence state allows EviRegionSearch to determine what evidence remains unresolved and use this information to guide subsequent navigation action selection.

\subsubsection{Necessity of Evidence-to-Action Translation.}
\label{sec:evi_region_search}

We next evaluate whether navigation actions should be adaptively selected according to the evolving evidence requirement. Specifically, we examine three aspects of action selection governing continued exploration: whether to redirect exploration to another region at the current magnification ({Move}), how to transition across magnifications ({Zoom In / Zoom Out}), and whether sufficient evidence has been acquired to terminate exploration ({Stop}). Each decision is evaluated through a controlled comparison while keeping the remaining navigation settings fixed.

\textbf{Move action.}
We first evaluate whether evidence-adaptive exploration improves QA performance over direct candidate sampling. To distinguish the contribution of the exploration strategy from candidate quality, we conduct comparisons under two candidate priors. Using random foreground candidates obtained with CLAM tissue segmentation~\cite{lu2021data}, we compare direct random sampling with exploration guided by EviRegionSearch. We then repeat the comparison using CrossMagLoc, where the baseline directly follows the localization ranking, while EviRegionSearch selects navigation actions according to the current question, accumulated evidence, m, and remaining evidence requirement. When Move is selected, it reuses the current heatmap to examine the highest-scoring unexamined region at the current magnification.

As shown in Table~\ref{results:navigation_ablation}, under the random candidate prior, question-conditioned region selection increases morphology GACC from 55.14\% to 68.00\% and PBR from 54.29\% to 74.29\%. With CrossMagLoc candidates, directly following the localization ranking achieves 70.00\% morphology GACC and 77.14\% PBR, whereas evidence-adaptive region selection further increases them to 76.00\% and 85.71\%, respectively. Overall GACC similarly increases from 47.15\% to 53.72\%.

These results show that EviRegionSearch \textbf{does not simply follow a fixed candidate ranking, but adaptively redirects exploration toward regions} relevant to the evidence currently required.

\textbf{Zoom In / Zoom Out actions.}
We next evaluate whether Zoom In / Zoom Out actions should be adaptively selected according to the remaining evidence requirement. We first compare the full setting with fixed-magnification exploration at 1.25$\times$, 2.5$\times$, 5$\times$, and 10$\times$, which removes magnification transitions entirely. We further evaluate a predefined coarse-to-fine sequence of 1.25$\times \rightarrow$ 2.5$\times \rightarrow$ 5$\times \rightarrow$ 10$\times$, which provides access to multiple magnifications but replaces evidence-adaptive action selection with a fixed transition order.

As shown in Table~\ref{results:navigation_ablation}, fixed-magnification exploration yields morphology GACC ranging from 33.14\% to 47.14\%, substantially below the full adaptive setting at 76.00\%. The predefined coarse-to-fine strategy improves morphology GACC to 65.71\% and overall GACC to 47.37\%, but remains below the full setting at 76.00\% and 53.72\%, respectively.

These results highlight the importance of \textbf{adaptively adjusting magnification} according to the remaining evidence requirement, rather than following a predefined coarse-to-fine traversal or remaining at a single magnification.

\textbf{Stop action.}
Finally, we evaluate whether exploration should terminate according to evidence sufficiency rather than a predefined navigation budget. In the full setting, EviRegionSearch selects Stop when the remaining evidence requirement has been resolved and the accumulated question-relevant evidence is sufficient for answering the current question. We compare this with a fixed-budget variant that terminates navigation after a predefined navigation budget while keeping the remaining navigation settings unchanged.

As shown in Table~\ref{results:navigation_ablation}, fixed-budget stopping reduces morphology GACC from 76.00\% to 49.71\% and PBR from 85.71\% to 48.57\%, while increasing IAR from 11.43\% to 37.14\%. Over all QA, GACC decreases from 53.72\% to 44.45\%, with PBR decreasing from 42.34\% to 32.85\%.

These results highlight the importance of \textbf{stopping according to evidence sufficiency}, with Stop selected when the remaining evidence requirement has been resolved rather than when a predefined navigation budget is reached.

Together, these controlled evaluations show that EviRegionSearch progressively maintains an evolving evidence state and uses it to adaptively determine where to explore, how to transition across magnifications, and when to terminate exploration.

\subsection{Qualitative Evaluation}
\label{sec:qualitative}

We further conduct qualitative analyses from two complementary perspectives. First, we examine the \textbf{interpretability of CrossMagLoc} through pathologist review of whether its localization heatmaps highlight diagnostically relevant regions across magnifications. Second, we visualize a \textbf{representative PathPilot workflow} to illustrate how the complete framework adaptively acquires and accumulates question-relevant evidence across regions and magnifications.

\subsubsection{Pathologist-in-the-loop Interpretability of CrossMagLoc.}
We assess the diagnostic relevance of CrossMagLoc-highlighted regions through independent pathologist review. Two pathologists, trained in Europe and Asia, respectively, review selected cases at the thumbnail, 1.25$\times$, 2.5$\times$, 5$\times$, and 10$\times$ levels, marking diagnostically informative regions and providing brief morphological explanations. Each pathologist repeats the assessment across multiple independent sessions under routine working conditions; regions consistently considered diagnostically relevant are retained as the final annotations. Fig.~\ref{fig5} presents these annotations and explanations alongside CrossMagLoc predictions and the original test-set viewing annotations.

CrossMagLoc broadly agrees with the original viewing annotations while prioritizing regions considered informative by the independent pathologists. At 2.5$\times$, it assigns greater attention to well-preserved, tumor-dense regions identified by both reviewers and lower attention to a region affected by severe crush artifact. The artifact-affected region remains partially informative for architectural assessment but is less suitable for evaluating cellular morphology. At higher magnifications, the highlighted regions exhibit diagnostically interpretable features, including retraction clefts between tumor nests and surrounding stroma, relatively uniform hyperchromatic tumor cells, and infiltrative tumor nests or cords.

The independent review also supports the \textbf{diagnostic relevance of additional regions} highlighted by CrossMagLoc. At 5$\times$, a high-attention region absent from the original viewing annotations is independently identified by the European pathologist as relevant for higher-magnification assessment. This example shows that differences from recorded viewing annotations can reflect additional diagnostic information, since a single viewing session captures only one possible diagnostic navigation pattern. Together, these observations support CrossMagLoc's ability to highlight diagnostically meaningful regions across magnifications beyond those recorded in the original annotations. Additional cases with pathologist annotations, morphological explanations, and CrossMagLoc heatmaps are provided in the Supplementary Material.

\subsubsection{Visualization of PathPilot Workflow.}

To qualitatively examine the navigation behavior of PathPilot, Fig.~\ref{fig6} visualizes a representative question-conditioned exploration trajectory. Starting from a thumbnail overview, the agent uses the heatmap generated by CrossMagLoc to obtain a diagnostically relevant starting ROI. EviRegionSearch then updates the accumulated question-relevant evidence and remaining evidence requirement, selecting subsequent navigation actions accordingly. This trajectory illustrates \textbf{evidence-adaptive exploration}: the agent examines diagnostically relevant regions, increases magnification when finer morphological evidence is needed, and stops once sufficient evidence has been accumulated to answer the question. Each ROI is represented as a $256\times256$ image, while its spatial footprint on the WSI varies with magnification, allowing examination of different fields of view. Additional workflows covering different navigation patterns and datasets are provided in the Supplementary Material.

\subsection{Ablation Studies.}
\label{sec:ablation}

We conduct a series of ablation studies to examine the two key components of PathPilot, CrossMagLoc and EviRegionSearch. For CrossMagLoc, we first evaluate whether question-agnostic diagnostic localization benefits subsequent exploration, followed by ablations of its supervision objective and architectural components for learning magnification-specific diagnostic localization. For EviRegionSearch, we examine Evidence Requirement Reasoning, which updates the accumulated question-relevant evidence and remaining evidence requirement, and Evidence-to-Action Translation, which uses these updated states to select the next navigation action. We further investigate the visual guidance strategy and the choice of LLM and VLM backbones in EviRegionSearch.

\subsubsection{CrossMagLoc.}

\textbf{Ablation Study of Diagnostic Localization.}
As discussed in Sec.~\ref{sec:evi_region_search} and shown in Table~\ref{results:navigation_ablation}, we evaluate the contribution of question-agnostic diagnostic localization learned from pathologist viewing behavior by comparing CrossMagLoc with random foreground sampling under the same EviRegionSearch settings. Using CrossMagLoc increases morphology GACC from 55.14\% to 70.00\% and PBR from 54.29\% to 77.14\%. These results demonstrate that \textbf{localizing diagnostically relevant regions supports more efficient subsequent exploration}.

\textbf{Ablation Study of CrossMagLoc Loss and Components via Multiple Instance Learning (MIL).}
To evaluate CrossMagLoc's loss and components independently of EviRegionSearch, we use four-class MIL classification to assess whether its heatmaps prioritize diagnostically informative regions (Table~\ref{table8}). We use the same Skin Eye-Tracking dataset to avoid introducing additional cross-dataset domain shift. At 10$\times$, patches are ranked by their diagnostic relevance heatmap scores, and the top 20\%, 40\%, 60\%, and 80\% are selected as input to ABMIL~\cite{ilse2018attention}. Classification performance is measured by AUC using 5-fold cross-validation with a 60\%:20\%:20\% train:validation:test split.

For the CrossMagLoc loss, the complete NDSL consistently achieves the \textbf{highest AUC across all patch budgets}. Removing weighted $\ell_1$ causes the largest average degradation across patch budgets, with AUC drops of 1.23–3.02 \%, while removing soft Dice or soft focal loss generally leads to smaller decreases. These results support the primary role of weighted $\ell_1$ supervision and the complementary contributions of soft Dice and soft focal loss in learning from sparse and potentially incomplete navigation annotations.

We further ablate the architectural components of CrossMagLoc. For this analysis, CMCF is divided into two branches: CMCF-L pairs the lower-magnification token with a downsampled version of the current representation, while CMCF-H pairs the higher-magnification token with an upsampled version, simulating adjacent magnification scales. Combining IMCA, CMCF-L, and CMCF-H achieves the highest AUC across all patch budgets, reaching 93.89\%, 93.29\%, 94.17\%, and 93.90\% using the top 20\%, 40\%, 60\%, and 80\% of patches, respectively. Removing either CMCF-L or CMCF-H consistently degrades performance, supporting their complementary contributions to magnification-specific diagnostic localization. Notably, achieving 93.89\% AUC with \textbf{only the top 20\% of patches} demonstrates that CrossMagLoc prioritizes diagnostically informative regions for downstream analysis.

\subsubsection{EviRegionSearch.}

\textbf{Ablation Study of Navigation Strategy.}
As discussed in Sec.~\ref{sec:eviregionsearch_eval} and summarized in Table~\ref{results:navigation_ablation}, we systematically ablate the main navigation mechanisms of EviRegionSearch, including question-conditioned region search, evidence requirement reasoning, adaptive magnification selection, and adaptive stopping. The corresponding controlled variants demonstrate that \textbf{each mechanism contributes to evidence-adaptive exploration across regions and magnifications}, with the complete EviRegionSearch configuration achieving the strongest overall performance.

\textbf{Ablation Study of Visual Guidance.}
We further evaluate the visual guidance $\tilde{Q}$ introduced in Sec.~3.3 by comparing three VLM prompting strategies while keeping all other settings fixed: no question information, directly providing the original question $Q$, and providing the transformed visual guidance $\tilde{Q}$. As shown in Table~\ref{results:backbone_ablation}, morphology GACC increases from 62.86\% without question information to 74.86\% when the original question is provided, and further to 76.00\% with visual guidance. The proposed guidance also achieves the highest morphology PBR of 85.71\% and the lowest IAR of 11.43\%. A consistent improvement is observed in the supporting average across all questions, where GACC increases from 41.39\% to 45.77\% and then to 53.72\% across the three settings, while IAR decreases to 20.44\% with visual guidance.

These results indicate that incorporating question-specific information is important for obtaining relevant ROI observations, while transforming the question into compact visual guidance \textbf{question-focused morphological observation} for subsequent evidence accumulation. We therefore use $\tilde{Q}$ as the VLM prompting strategy in PathPilot.

\textbf{Ablation Study of LLM and VLM Backbones.}
We evaluate different LLM and VLM backbones while keeping the remaining components fixed (Table~\ref{results:backbone_ablation}). With Patho-R1-7B as the VLM, we compare Qwen2.5, Qwen3, Phi-4~\cite{abdin2024phi}, DeepSeek-R1-Distill-Qwen-14B~\cite{guo2025deepseek}, and Mistral-Small-3.2-24B~\cite{mistralsmall32}. Qwen3-14B achieves the \textbf{highest morphology and overall GACC}, reaching 76.00\% and 53.72\%, respectively. Larger models do not consistently improve performance: Qwen3-32B and Qwen2.5-32B achieve morphology GACC of 67.14\% and 69.71\%, respectively. The other LLM families also support the framework, indicating compatibility across model architectures.

With Qwen3-14B fixed, we compare HealthGPT-Pro-8B, Quilt-LLaVA, MedGemma-1.5-4B, and Patho-R1-7B as VLM backbones. Patho-R1-7B achieves the \textbf{strongest overall performance}, while HealthGPT-Pro-8B and Quilt-LLaVA remain competitive on morphology questions, with GACC of 75.14\% and 72.86\%, but perform less well across all questions. These results support the use of different VLM backbones for ROI-level morphological observation. We adopt Qwen3-14B and Patho-R1-7B in the final configuration.

\section{Discussion}
\label{sec:discussion}

Regarding the evaluation of PathPilot, unlike conventional tasks, e.g., tumor segmentation or detection, there is no unique ground-truth navigation behavior. A slide may contain multiple spatially separated tumor regions that provide equivalent diagnostic evidence; one pathologist may inspect one region in detail, while another may focus on a different region, yet both can reach the same correct diagnosis. Navigation behavior can also vary substantially with experience and individual practice: less experienced pathologists may scan more broadly, repeatedly revisit previously viewed regions, or move back and forth across magnifications before identifying the relevant evidence, whereas experienced pathologists may recognize diagnostically informative ROIs much more directly. Therefore, neither reproducing a specific viewing trajectory nor matching a single set of attended regions can fully characterize navigation quality. To address this inherent ambiguity, we evaluate PathPilot from multiple complementary perspectives, including downstream open-ended QA performance, alignment with pathologist region prioritization, coverage of annotated tumor regions, controlled ablations of adaptive navigation mechanisms, and independent pathologist assessment. Together, these evaluations assess whether PathPilot can identify sufficient and diagnostically meaningful evidence without assuming that a single human navigation pattern constitutes the only correct solution.

\section{Conclusions}
In this paper, we propose \textbf{PathPilot}, an adaptive question-conditioned Multi-Magnification WSI Navigation Agent that combines \textbf{CrossMagLoc} for magnification-specific diagnostic localization with \textbf{EviRegionSearch} for evidence-adaptive, question-conditioned exploration. Across five open-ended QA cohorts, PathPilot achieves the highest overall GACC, outperforming the strongest baselines by \textbf{8.46\%} and \textbf{9.03\%} on the internal and external cohorts, respectively. CrossMagLoc further improves region-ranking correlation from \textbf{0.56 to 0.70} while covering \textbf{85--100\%} of annotated tumor regions across magnifications, with ablation studies and independent pathologist assessment further supporting the effectiveness of the proposed navigation strategy. Compared with conventional WSI analysis based on predefined sampling at a fixed magnification, PathPilot adaptively explores gigapixel WSIs across regions and magnifications, offering \textbf{a new approach to adaptive and effective WSI analysis} in computational pathology.

\section{Data Availability}

The Skin Eye-Tracking dataset is available through \url{https://github.com/MasyerN/PEAN}. The TCGA-SKCM and TCGA-LIHC QA cohorts are derived from WSI-Bench, available through \url{https://github.com/XinhengLyu/WSI-LLaVA}. The TUM-Uteria dataset is available at \url{https://huggingface.co/datasets/SchuefflerLab/TUM_Uteria}, subject to acceptance of its Data Use Agreement. All other TUM datasets used in this study were collected in-house and cannot be shared. The source code for PathPilot will be made publicly available upon publication.

\section{Declaration of generative AI and AI-assisted technologies in the writing process}

During the preparation of this work the authors used ChatGPT in order to improve language and readability only. After using this tool/service, the authors reviewed and edited the content as needed and take full responsibility for the content.

\section{Ethics Statement}
This study was approved by the Ethics Committee of the Technical University of Munich (approval number 2023-665-1-S-CT) and was conducted in accordance with the Declaration of Helsinki. All data used in this study was anonymized prior to model training.

\section{Acknowledgments}
This work was supported by the BMFTR-funded U-PathoAgent project (031L0333), the IMI BIGPICTURE project (IMI945358), the SATURN3 project (01KD2206C), and the Bavarian AI Foundation Model Initiative (AI-Bay.eu), funded by the Bavarian State Ministry of Science and the Arts (StMWK).

\FloatBarrier

\bibliographystyle{model2-names.bst} \biboptions{authoryear}
\bibliography{egbib}

\end{document}